\documentclass[journal,comsoc,onecolumn]{IEEEtran}

\usepackage{lineno}
\usepackage{url}
\usepackage{graphicx}
\usepackage{caption}
\usepackage{subcaption}
\usepackage{epstopdf}
\usepackage{amsmath}
\usepackage{comment}
\usepackage{array}
\usepackage{textgreek}
\usepackage{pbox}
\usepackage{multirow}
\usepackage{todonotes}
\usepackage{floatrow}
\usepackage{tabularx}
\usepackage{makecell}
\usepackage{amssymb} % For symbols like \mathbb
\usepackage{amsfonts} 

\usepackage{bm,graphicx}
\usepackage{amsmath}
\usepackage[inline, shortlabels]{enumitem}

\usepackage{booktabs} % For formal tables
\usepackage{soul}
\usepackage{color, colortbl}
\usepackage{mathtools}
\usepackage{rotating}
\definecolor{Gray}{gray}{0.9}
\definecolor{darkgray}{rgb}{0.66, 0.66, 0.66}

\begin{document}

\title{IMSAHLO: Integrating Multi-Scale Attention and Hybrid Loss Optimization Framework for Robust Neuronal Brain Cell Segmentation
}
%\author{Rimjhim,~\IEEEmembership{Student Member,~IEEE,}

\author{
    \IEEEauthorblockN{
        Ujjwal Jain\IEEEauthorrefmark{1},
        Oshin Misra\IEEEauthorrefmark{1},
        Roshni Chakraborty\IEEEauthorrefmark{1},
        Mahua Bhattacharya\IEEEauthorrefmark{1}
    }\\
    \IEEEauthorblockA{
        \IEEEauthorrefmark{1}ABV-IIITM Gwalior, India \\
        Emails: imt\_2021105@iiitm.ac.in, oshin@iiitm.ac.in, roshni@iiitm.ac.in, mb@iiitm.ac.in
    }
}

\iffalse
\author{Ujjwal Jain, Oshin Misra, Roshni Chakraborty, and Mahua Bhattacharya}
\fi

%

% \IEEEtitleabstractindextext{%

\maketitle

\begin{abstract}
In the era of AI and advanced technology, accurate segmentation is possible to solve major types of cell segmentation challenges.
On the same page for performing quantitative morphology analysis, different model architectures are used for different types of neuronal cell datasets. In the case of the heterogeneous mixture of dense cell clusters and sparse, low-contrast areas, it is critical to process them comparatively. Using fluorescence microscopy images is a fundamental task for quantitative analysis in computational neuroscience, although it is impeded by significant challenges, including the coexistence of densely packed and sparsely distributed cells, complex and overlapping morphologies, and severe class imbalance. Conventional deep learning models often fail to preserve fine topological details and accurately delineate boundaries in these challenging conditions. To address these limitations, we propose a novel deep learning framework, IMSAHLO (Integrating Multi-Scale Attention and Hybrid Loss Optimization), for robust and adaptive neuronal segmentation. The core of our model features Multi-Scale Dense Blocks (MSDBs) to capture features at various receptive fields, effectively handling variations in cell density, and a Hierarchical Attention (HA) mechanism that adaptively focuses on salient morphological features to preserve Region of Interest (ROI) boundary details. Furthermore, we introduce a novel hybrid loss function that synergistically combines Tversky and Focal loss to combat class imbalance, with a topology-aware Centerline Dice (clDice) loss and a Contour-Weighted Boundary loss to ensure topological continuity and precise separation of adjacent cells. Large-scale experiments on the public Fluorescent Neuronal Cells (FNC) dataset illustrate our framework outperforms state-of-the-art architectures considerably, reporting better precision (81.4\%), macro F1 score (82.7\%), micro F1 score (83.3\%) and balanced accuracy (99.5\%) on difficult dense and sparse cases. Ablation studies validate individual and synergistic benefits of multi-scale attention and hybrid loss terms. The work establishes a foundation for generalizable segmentation models that can be used with a wide range of biomedical imaging modalities and pushes AI-assisted analysis to high-throughput neurobiological pipelines.
\end{abstract}

% }
\maketitle

\IEEEpeerreviewmaketitle

\section{Introduction} \label{s:intro}
\par 
In computational cellular neuroscience, accurate segmentation of neuronal cells structure from fluorescence microscopy images is a prerequisite for several downstream tasks, such as, neurite tracing \cite{nourollah2024quantifying}, synapse identification \cite{zarb2025multimodal}, morphological classification \cite{debeuckeleer2025unbiased}, and quantitative neurobiology applications \cite{bjerke2023scaling}. Advanced imaging techniques, such as fluorescent labelling or confocal microscopy, have made it possible to view fine-grained morphological features of neuronal brain cells \cite{Krikid2024StateofthetotDL}. 
The fundamental challenges are mainly due to the nature of neuronal microscopy data, i.e., heterogeneous mix of densely covered regions of cells with superimposed processes and sparsely populated areas where faint structures are easily lost in background noise. This is further complicated by low signal-to-noise (SNR) ratios, non-uniform distribution of fluorescence across the sample, and poorly-defined or weak contrast boundaries \cite{Zhu2022A_Compound_Loss_Function}. Moreover, these images are plagued by a strong class imbalance of the foreground and background, since cell structures only occupy a few pixels, which are hard to be learned for conventional segmentation pipelines \cite{Zhu2022A_Compound_Loss_Function, Yeung2022UnifiedFL}.

% \textbf{well written, every point should be cited}
% \textbf{Is this the first work?}

\par Recent advances in deep learning based models have essentially led to efficient segmentation frameworks that outperform primitive thresholding or region-growth methods\cite{haque2020deep, xu2024advances}. The U-Net architecture \cite{ronneberger2015u} and its variations \cite{zhou2018unet++,Oktay2018AttentionUL,Chen2021TransunetTM,Wu2021UNetCW,Valanarasu2022UNEXTMR,Ma2024DTASUNetAL,peng2025u}, have become the most preferred biomedical segmentation backbones. The original U-Net's power stems from its symmetric encoder-decoder setup with skip connections, which enables precise localization by combining deep, semantic features with shallow, fine-grained spatial information. Building on this principle, its many variations seek to enhance performance by redesigning skip pathways or incorporating attention mechanisms. For example, the novelty of UNet++ \cite{zhou2018unet++} that includes a layered dense skip connection \cite{peng2025u} further narrowed the semantic gap between encoder and decoder attributes while also enhancing multi-scale learning\cite{chang2019multi}. With recalibration at spatial and channel-specific levels, Attention-based models like Attention U-Net \cite{Oktay2018AttentionUL} and ECA-Net \cite{Wang2020ECANetEC} ensure improved localisation. Also, Squeeze-and-Excitation(SE) modules \cite{hu2018squeeze} and Multi-scale attention mechanisms \cite{Wu2021UNetCW, huang2017multi} showed much better performances than prior methods in maintaining fine-grained structural features. However, these sophisticated modules can introduce significant computational overhead and may still struggle to differentiate the faintest neuronal structures from complex background noise.
% \textbf{if provide give one sentcne insights} \textbf{where did this lack}

\par Alongside these architectural advancements, the selection of the loss function is equally significant in architectural design \cite{Jadon2020ASO}. Conventional losses, such as the Dice \cite{Sudre2017GeneralisedDO} or Binary Cross-Entropy \cite{Yeung2022UnifiedFL}, operate on the principle of region overlap and leave out consideration of topological or boundary-specific attributes that are crucial for neuronal morphology \cite{Zhu2022A_Compound_Loss_Function,Zhou2025BoundaryawareAC,Clissa2023OptimizingDL}. While recent works have proposed specialized loss functions—Tversky \cite{Salehi2017TverskyLF,Abraham2019ANF}, Contour-aware \cite{Chen2020ContourawareLB,Huang2024ContourweightedLF}, clDice \cite{Shi2024CenterlineBD}, and SPW Loss \cite{Lu2025SteerablePW}—that emphasize the region-boundary trade-off and centerline continuity, they often excel at one objective at the expense of others. For example, a strong focus on boundary definition might not guarantee the topological continuity of fine neuron cells. This creates a critical gap, as no single existing loss function comprehensively balances the competing demands of managing severe class imbalance, ensuring boundary precision, and preserving the topological integrity of complex neuronal structures. This limitation motivates the need for a carefully designed hybrid loss function that can provide multifaceted supervision.

\par In this paper, we propose IMSAHLO (Integrate Multi-Scale Attention Hybrid Loss Optimization), a comprehensive segmentation framework specifically engineered to address these multifaceted challenges of neuronal segmentation. A summary of the main contributions are : 

% \textbf{not discussed}

\begin{enumerate}
    \item A novel deep learning architecture which integrates Multi-Scale Dense Blocks to address morphological variability and Hierarchical Attention mechanism to dynamically concentrate on informative regions, both dense and sparse.
    \item A hybrid loss function that is carefully crafted to combine Tversky, Focal, Contour-Weighted, CenterlineDice (clDice) losses to offer a multi-faceted supervisory signal that optimally weights vital aspects, including recall, boundary precision, class imbalance, and topological continuity.
    \item Extensive experiments and ablation studies on the publicly available Fluorescent Neuronal Cells (FNC) dataset \cite{Morelli2021AutomatingCC}. As detailed in Table~\ref {tab:baseline_models} and Table~\ref {tab:loss_ablation}, our results show that IMSAHLO establishes a new state-of-the-art across all evaluation criteria, achieving a Macro F1 of 0.827, Precision of 0.814, and Recall of 0.789. Critically, IMSAHLO achieves a Dice Similarity Coefficient (DSC) of 0.801 and an Intersection over Union (IoU) of 0.784, representing a significant 9.7\% improvement in Dice score over the next-best performing method.
    % \textbf{add more quantified results with details on comparison}
\end{enumerate}
This work provides a foundation for building segmentation models that can be applied across a wide range of biomedical images. More than just a performance boost, our approach can directly accelerate high-throughput neurobiological research. For instance, it can be used to automate critical tasks like large-scale cell tracing, cell counting in drug screening assays, or the morphological classification of thousands of cells to study disease phenotypes. Rest of the paper is organized as follows. We discuss related works in Section~\ref{s:rworks}. In Section~\ref{s:pstat}, we present problem definition and discuss details of the proposed approach in Section~\ref{s:prop}. We discuss the experiment details in Section~\ref{s:expt}, results in Section~\ref{s:res} and finally, conclusions in Section~\ref{s:con}.
% \textbf{examples, dont give generic broad claims without specific examples, it goes against}.

\section{Related Works} \label{s:rworks}
% \textbf{which two streams?}
% \textbf{write a small sentence or two, this is a very long sentence}. 
%  \textbf{do not write in bullet points for related works, write in paragraph style and each paper does not have to discussed in so much detail, only the main idea and the failure analysis. Papers should be divided into groups}\\
Advances in deep neural networks and tailored loss functions have significantly improved cell segmentation in microscopy, especially for complex and noisy visual data. We provide a chronological summary of these two streams based on deep learning approaches and loss function analysis, relevant to the challenges of segmenting morphologically complex structures in dense and sparse fluorescence imaging environments. \par
Based on Deep Learning approaches, neural architectures and attention-driven mechanisms have greatly facilitated the development of cell segmentation in microscopy\cite{Krikid2024StateofthetotDL}\cite{vaswani2017attention}\cite{soydaner2022attention}. These trends are in response to a rising need to extract fine-grained, morphologically relevant features from noisy and class-imbalanced biological data. Depending on CNN-based encoder-decoder segmentation\cite{Hiramatsu2018CellIS}, U-Net \cite{ronneberger2015u} revolutionized the field of study through an encoder–decoder architecture connected with skip links to retrieve spatial information. By utilizing dense nested skip pathways, its successor U-Net++ \cite{zhou2018unet++} improved the multi-scale feature fusion and reduced the semantic gap between encoder and decoder enhancing the model's ability to segment objects of varying sizes. Squeeze-and-Excitation (SE) blocks \cite{hu2018squeeze} that enabled channel-wise recalibration \cite{roy2018recalibrating}, and Multi-Scale Dense (MSD) networks \cite{huang2017multi} that were capable of learning wide receptive fields without being too deep were two other convolutional developments around that time. DeepLabV3+ \cite{Chen2018EncoderdecoderWA} significantly improved spatial sensitivity with atrous separable convolutions for dense predictions. Lightweight and mobile-friendly design that combined transformer-inspired global reasoning with convolutional efficiency has been explored by MobileViT \cite{Mehta2021MobilevitLG}.\\
 Motivated by the "Attention is All You Need" concept \cite{vaswani2017attention}, the attention mechanism has been introduced into medical image segmentation for spatial attention to handle low-contrast and noisy signals, particularly in sparse regions. Attention U-Net \cite{Oktay2018AttentionUL} employs a soft attention gating to emphasize meaningful spatial features in biological instances. ECA-Net \cite{Wang2020ECANetEC} contributes a relatively efficient channel attention, and U-Net variants with multi-scale attention also show superiority in organ segmentation (e.g., liver) \cite{Wu2021UNetCW}. Meanwhile, dual attention branches have been explored in other models \cite{Wang2024MultimodalPA}, and a hierarchical attention architecture \cite{Ding2019HierarchicalAN} was developed to improve fine-grained localisation of cellular and subcellular components. Residual and channel-prior attention mechanisms were also investigated in medical situations \cite{Xie2023ResidualTW} \cite{Huang2024ChannelPC}, which focus on better gradient flow and structure relevance. These approaches demonstrated promising improvements in delineating extended, low-contrast anatomical boundaries observed in fluorescence microscopy.\\
 Transformer-based reasoning has allowed U-shaped systems to obtain the next-generation improvements in segmentation. TransUNet \cite{Chen2021TransunetTM} merged ViT-based global encoding with convolutional decoders and reached state-of-the-art performance in a wide range of segmentation tasks. DTASUNet \cite{Ma2024DTASUNetAL} improved the paradigm by the integration of a dual transformer branch, i.e., one local and one global, in addition to an attention-supervised decoder, and obtained competitive results in brain tumour segmentation. Moreover, UNeXt \cite{Valanarasu2022UNEXTMR} and residual-transformer hybrids \cite{Xie2023ResidualTW} have been suggested to be as efficient MLP- or transformer-based U-Net substitutes with competitive performance and low-cost computation. These models are especially appealing for dense biological image analysis because of their hierarchical feature representation.\\
Recent investigations, such as Zhang et al. \cite{Zhang2022SemanticSM}, which have specifically tackled the problem of neural segmentation in the low data regime by leveraging semantic priors in a few-shot setting. Domain-generalized methods such as multimodal attention fusion \cite{Wang2024MultimodalPA} and cross-domain augmentation \cite{Xu2024CrossDomainAG} have also increased generalization to different imaging modalities and staining techniques. Although early U-Net-like models have led to a reliable platform for biomedical cell segmentation, current models that integrate attention and transformer-based modules have achieved better results in the challenging cell segmentation problem. Nevertheless, the trade-off between structural-detail preservation and efficiency remains a problem, especially in high-resolution fluorescence microscopy settings. Our proposed architecture, IMSAHLO, expands on the above foundations by combining Multi-Scale Dense blocks and Hierarchical Attention into a cohesive and computationally efficient model designed specifically for neuronal cell segmentation.

Based on Loss Function Analysis, segmentation of microscopic images, especially neuronal cells, is often hindered by class imbalance, faint boundaries, and overlapping structures \cite{Zhu2022A_Compound_Loss_Function}. To overcome them, the community has incrementally designed a variety of ad-hoc loss functions \cite{Clissa2023OptimizingDL} \cite{Terven2023LossFA}. These losses can generally be grouped according to the concerned aspect: region-level accuracy, boundary sensitivity, structural continuity, and hybrid objectives \cite{Jadon2020ASO}.
For defining foundational Region-Aware Losses, Salehi et al. proposed Tversky Loss \cite{Salehi2017TverskyLF}, a generalization of the Dice coefficient that can adjust the tradeoff between false positives and false negatives by adjusting weights α and β. This technique was particularly beneficial for class-imbalanced biomedical segmentation. Lin et al. \cite{lin2017focal} proposed Focal Loss, a method that rescales the contribution of difficult-to-classify pixels to down-weight their presence compared to background pixels and hence improve the performance on under-represented structures. Sudre et al. applied Generalised Dice Loss \cite{Sudre2017GeneralisedDO} in this stage, which scales class weights by the inverse volume to favour minor classes during the training procedure. \\
After that, for boundary awareness and hybridization, authors re-emphasized the need for edge-preserving segmentation. Abraham and Khan \cite{Abraham2019ANF} introduced the focused Tversky Loss that combines region sensitivity with emphasis on problematic pixels using a focused modulation term. Kervadec et al. \cite{Kervadec2019BoundaryLF} introduced Boundary Loss, which directly penalised misalignment at the boundaries of segmentations and worked particularly well on datasets with small foreground objects. Contour-Aware Loss \cite{Chen2020ContourawareLB} and Hybrid-Loss Supervision \cite{Cheng2020HybridlossSF} are intended to integrate region-based and edge-based objectives, leading to improved pixel-wise accuracy and structural correspondence. Jadon \cite{Jadon2020ASO} compared these functions, with guidance on their trade-offs and how to use them within a practical segue to segment a real-world segmentation pipeline.\\
Structural and Shape-Aware objectives shifted the focus on modelling spatial continuity and irregular morphologies. Yeung et al. \cite{Yeung2022UnifiedFL} presented Unified Focal Loss (UFL), which combines dice and cross-entropy to keep both merits. Zhu et al. \cite{Zhu2022A_Compound_Loss_Function} produced a Shape-Aware Compound Loss that is suited to overlapping or unevenly shaped cells, common complications in fluorescence microscopy. Guo et al. \cite{Guo2022ContourLF} presented Distance-Transform-based Contour Loss, which is capable of better fitting borders by suppressing the differences of the k-step distance transform. Clissa et al. \cite{Clissa2023OptimizingDL} performed an extensive benchmarking study and showed that hybrid losses mixing boundary and regional cues resulted in increased generalization in fluorescence microscopy. Terven et al. \cite{Terven2023LossFA} also studied different loss functions and evaluation metrics in order to standardize the performance interpretation in challenging segmentation scenarios.\\
Recent works focus on structure-aware, modality-agnostic supervision. Huang and Sui \cite{Huang2024ContourweightedLF} proposed the Contour-Weighted Loss, which assigns dynamic weights to the boundary areas in sparse and crowded images to improve the segmentation accuracy of fine-scale features. Shi et al. \cite{Shi2024CenterlineBD} introduced the Centerline Boundary Dice Loss, which tried to preserve the continuity of elongated structures as much as possible by imposing the centerline completeness. Zhou \cite{Zhou2025BoundaryawareAC} extended this with Boundary-Aware Cross Modality Loss, which efficiently handles segmentation on multi-modal medical images (e.g., MRI + CT). Steerable Pyramid Weighted Loss was suggested by Lu \cite{Lu2025SteerablePW}, which, with the help of a scale-aware filter, preserves the delicate patterns under high-resolution down-sampling limitations.\\
From region-balancing to structure-preserving, these evolving loss functions suggest an increasingly better understanding of the difficulties in biomedical segmentation. Hybrid losses balancing region, contour, and continuity objectives have been found to work well for fluorescent neuronal microscopy, in which neurons come in different sizes, intensities, and spatial distributions. In particular, our proposed method follows this study by combining Tversky, Contour-Weighted, Focal, and Centerline Dice terms into a unified hybrid loss function. This allows precise tracing of dense clusters and thin processes with high topological truth and generalization ability in various imaging conditions.

\section{Problem Statement} \label{s:pstat}
\par 
While powerful deep learning architectures, including U-Net++ \cite{zhou2018unet++}, attention-based models \cite{Oktay2018AttentionUL}, and transformer variants \cite{Chen2021TransunetTM, Mehta2021MobilevitLG, Ma2024DTASUNetAL} have significantly advanced neuronal segmentation, their performance can be inconsistent when faced with the full spectrum of challenges in fluorescence microscopy. Although these models often excel at particular aspects of segmentation, they tend to struggle when required to handle several competing data complexities simultaneously. This performance gap arises because they were not explicitly designed to address the four main challenges listed below concurrently:

% \textbf{you can not make a stratemet where it seems your proposed approach solves everyting that others do not}
% \textbf{generic statement again and goes against your introduction}
\begin{enumerate}
    \item \textit{Morphological and Distributional Heterogeneity:} A single framework has to perform well on both of the conflicting cases, i.e., the densely clustered cases which have a lot of cell overlap and adhesion and have ambiguous boundaries, and the sparsely distributed cases with faint signals and severe foreground vs. background class imbalance, causing missed detections.

    \item \textit{Preservation of Fine-Grained Topology:} The continuity of thin elongate cell processes is an important issue in quantitative neurobiology. Standard segmentation processes may lead to fragmented (discontinuous) formatting of these structures, which may hinder the conduct of downstream analyses, including connectivity mapping.

    \item \textit{Boundary Ambiguity and Low Contrast:} The model should be able to accurately trace the cell boundary in case of varying signal-to-noise ratios, nonidentical fluorescent staining in each cell, and the intervening separation between neighboring cells and background. 
  
    \item \textit{Lack of Integrated Architecture-Loss Co-Design:} The bottleneck of the aforementioned approaches derives from the lack of a unified segmentation method to co-architecture optimize loss. A principal solution would be to learn an architecture capable of learning multi-scale discriminative features with explicit supervision on boundaries, topology, and class imbalance jointly, generalized for all domains.
\end{enumerate}
%\vspace{4pt}
\subsection{Problem Definition}
To address these challenges, we formalize the task of neuronal segmentation as a joint optimization problem. Let $I \in \mathbb{R}^{H \times W \times C}$ be an input microscopy image from the domain $\mathcal{I}$, and let $Y \in \{0, 1\}^{H \times W \times K}$ be the corresponding ground-truth segmentation map for $K$ classes (e.g., $K=2$ for foreground and background).

Our goal is to learn a mapping function $f_\theta: \mathcal{I} \to [0, 1]^{H \times W \times K}$, parameterized by $\theta$ (i.e., the neural network weights), which produces a predicted probability map $P = f_\theta(I)$. The optimal parameters $\theta^*$ are found by minimizing a loss function $\mathcal{L}$ over a training dataset $\mathcal{D} = \{(I_n, Y_n)\}_{n=1}^N$:

\begin{equation}
    \theta^* = \arg\min_{\theta} \frac{1}{N} \sum_{n=1}^{N} \mathcal{L}\big(f_{\theta}(I_n), Y_n\big)
    \label{eq:optimal_theta}
\end{equation}

Given the multifaceted challenges identified, a standard segmentation loss (e.g., categorical cross-entropy or a simple Dice score) is insufficient as it fails to capture the competing objectives. Our core hypothesis, which \textit{directly addresses Lack of Integrated Architecture-Loss Co-Design (Challenge 4)}, is that a robust solution requires a compound objective function that explicitly and simultaneously penalizes errors in region, boundary, and topology. We therefore define the problem as the minimization of a compound loss \begin{align}
    \mathcal{L}_{\text{total}}(P, Y) 
    &= \lambda_1 \mathcal{L}_{\text{region}}(P, Y) \nonumber \\
    &\quad + \lambda_2 \mathcal{L}_{\text{boundary}}(P, Y) 
    + \lambda_3 \mathcal{L}_{\text{topology}}(P, Y)
    \label{eq:total_loss}
\end{align}
where:
\begin{itemize}
    \item $\mathcal{L}_{\text{region}}$ is a region-based loss function (e.g., a variant of Tversky or Focal loss) designed to handle \textit{Morphological and Distributional Heterogeneity (Challenge 1)} by mitigating class imbalance.
     \item $\mathcal{L}_{\text{topology}}$ is a topology-aware loss (e.g., a loss based on persistent homology or a connectivity-based regularizer) designed to ensure the \textit{Preservation of Fine-Grained Topology (Challenge 2)}.
    \item $\lambda_1, \lambda_2, \lambda_3$ are weighting hyper-parameters that balance the contribution of each competing objective.
    \item $\mathcal{L}_{\text{boundary}}$ is a boundary-focused loss (e.g., a weighted Boundary or Hausdorff-based loss) that applies explicit supervision to pixel interfaces, thereby addressing \textit{Boundary Ambiguity and Low Contrast (Challenge 3)}.
   
\end{itemize}

Motivated by this formulation, we propose the IMSAHLO framework. It is a hybrid architecture-loss approach that materializes this objective, integrating multi-scale attention and dense encoding (the \textit{architecture}, $f_\theta$) with a novel instantiation of the compound loss $\mathcal{L}_{\text{total}}$ (the \textit{loss}) to achieve topology-aware segmentation of challenging neuronal microscopy images. We discuss the Proposed framework in detail next.

\begin{figure*}[t]
    \centering
    \includegraphics[width=1\linewidth]{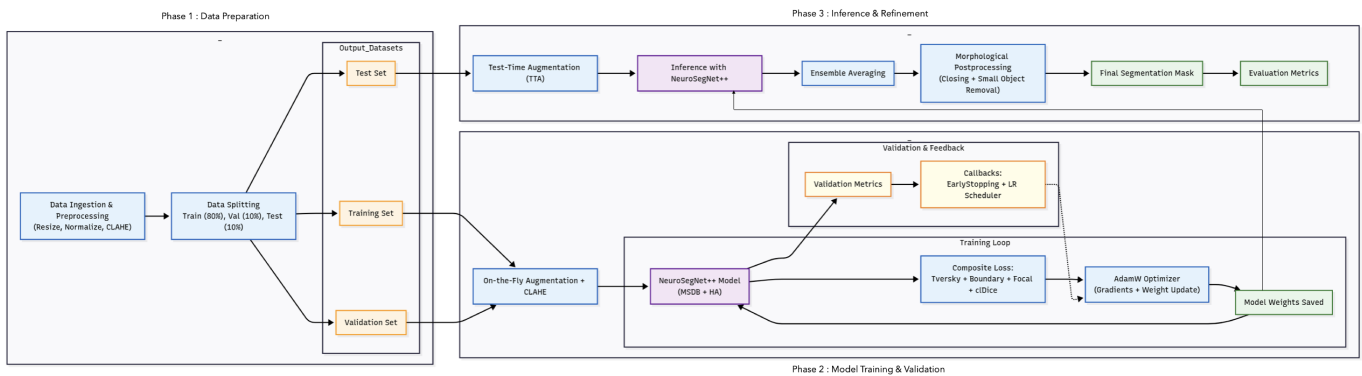}
    \caption{End-to-End Pipeline Architecture of IMSAHLO for
Fluorescent Neuronal Cell Segmentation}
    \label{fig:pipeline_arch}
\end{figure*}

\section{Proposed Approach}\label{s:prop}
\begin{figure*}[t]
    \centering
    \includegraphics[width=1\linewidth]{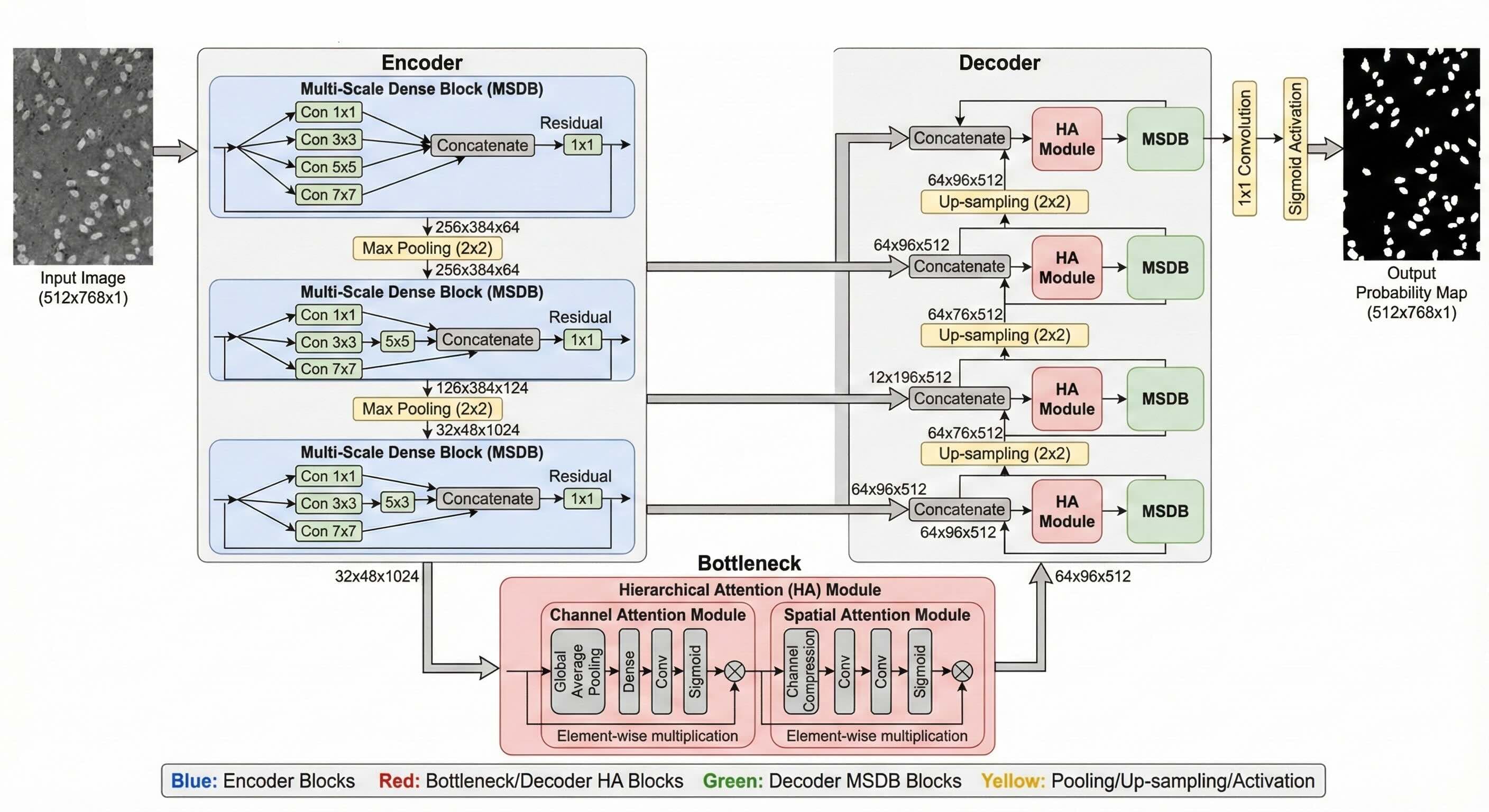}
    \caption{IMSAHLO: Detailed Architecture for Biomedical Image Segmentation}
    \label{fig:imsahlo_arch}
\end{figure*}

Segmenting fluorescently-labeled neuronal cells in microscopy images is a foundational task in connectomics, neuroscience, and cell biology. To overcome the inherent challenges in this task, we developed IMSAHLO, a deep learning system that integrates a tailored network architecture with a multi-component loss function specifically tuned for neuronal morphology. The IMSAHLO architecture, shown in Figure \ref{fig:imsahlo_arch}, directly targets the four main challenges of neuronal segmentation. To handle morphological and distributional heterogeneity, we employ Multi-Scale Dense Blocks (MSDB) that learn features across diverse receptive fields. To address boundary ambiguity and preserve fine-grained topology, we introduce a Hierarchical Attention (HA) mechanism that focuses on meaningful details, complemented by a novel hybrid loss function that supervises both edge sharpness and structural coherence. Finally, we demonstrate the power of this integrated architecture-loss co-design by analyzing how these components synergize to achieve accurate cell segmentation. The end-to-end pipeline, illustrated in Figure \ref{fig:pipeline_arch}, consists of three main phases : Data Preparation, System Architecture, and Inference and Refinement, which we discuss in detail next.
% \textbf{RC I remember discussing these names, we cant keep these names}
\subsection{Phase-I: Data Preparation}
  \textbf{Data Preprocessing and Splitting} : 
  This study utilized the Fluorescent Neuronal Cells Dataset \footnote{https://www.kaggle.com/datasets/nbroad/fluorescent-neuronal-cells}, which consists of grayscale microscopy images and corresponding binary masks. These images capture high-resolution views of mouse brain slices, providing a challenging benchmark for segmenting intricate neuronal structures. The preprocessing pipeline was designed to balance the preservation of fine morphological details with computational feasibility. All images and masks were uniformly resized to a spatial resolution of 512×768 pixels using inter-area interpolation. Finally, a singleton channel dimension was added to format the input tensors into a shape of (512,768,1), conforming to the standard input requirements for 2D convolutional networks. Visual inspection confirmed the integrity and alignment of the preprocessed data. To ensure a robust evaluation of the model's generalization capabilities and to mitigate the risk of data leakage from spatial and morphological similarities inherent in microscopy images, the dataset was partitioned into training, validation, and testing sets. A stratified sampling strategy was employed to ensure that each split contained a proportional representation of the data's diversity, including features such as dense cell clusters and isolated neurons. The reproducibility of these data partitions was guaranteed through the use of a fixed random state.
    
    \textbf{Data Augmentation:}
    To enhance the model's robustness and generalization capabilities, a comprehensive data augmentation pipeline was implemented. This strategy served as a crucial regularization mechanism, exposing the model to a diverse set of transformations that simulate the wide range of variations and artifacts common in biomedical imaging, such as those arising from sample preparation or hardware inconsistencies. The pipeline comprised three categories of transformations, which we discuss next.
    \begin{enumerate}
        \item \textit{Geometric Augmentations:} To achieve invariance to spatial transformations, a series of geometric augmentations were applied. These included elastic and grid distortions to model plausible non-linear tissue warping, alongside rigid transformations (flips) and affine transformations (rotations up to ±45 degree and random cropping). This ensured the model remained robust to changes in cellular orientation and the imaging field-of-view.
        
         \item \textit{Photometric Augmentations:} To improve robustness against variations in image quality and intensity, photometric transformations were employed. Random brightness and contrast adjustments (up to ±30\%) were used to simulate inconsistent illumination or fluorescent labeling. Furthermore, Gaussian noise and blur were introduced to account for sensor noise and optical effects encountered during image acquisition.
         \item \textit{CLAHE-Driven Local Contrast Enhancement:} To mitigate issues arising from low local contrast, a common artifact in images of weakly fluorescent cells, Contrast Limited Adaptive Histogram Equalization (CLAHE) \cite{mohammed2025contrast} was applied as a preprocessing step. This technique enhances contrast by performing histogram equalization on small, overlapping image tiles. A tile size of 8×8 pixels was selected to balance the enhancement of fine details and larger structures, while a clip limit of 3.0 was used to prevent the over-amplification of noise. This approach effectively emphasizes faint cell borders and augments mid-intensity boundaries without saturating bright regions, which is critical for discriminating fine neuronal projections from background fluorescence.
          \item \textit{Data Generator for Augmentation-Aware Training:} This generator performs online data augmentation and applies CLAHE normalization on-the-fly, which enhances memory efficiency and training speed by processing data in batches. Unlike static preprocessing, this method generates a virtually infinite stream of unique training samples, providing a powerful regularization effect. Crucially, all geometric transformations were applied synchronously to both the images and their corresponding masks to maintain precise structural alignment. Following the augmentation step, CLAHE was reapplied to ensure that contrast remained consistently enhanced, even after transformations that could alter pixel intensity distributions.
    \end{enumerate}

    % \textbf{very lengthy sentence}
    % \textbf{these last two sentences should come before and focus in small concise clear sentences, then write what you do. its not the other way around}.

\subsection{Phase-II: System Architecture}
    \begin{figure*}[t]
        \centering
        \includegraphics[width=1\linewidth]{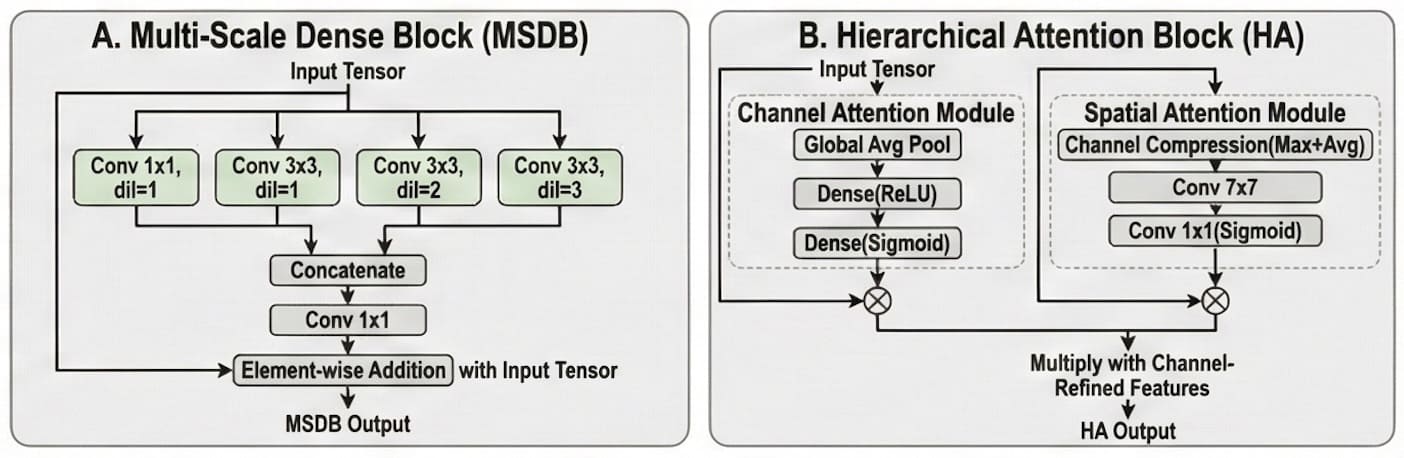}
        \caption{(A) Multi-Scale Attention Block, (B) Hierarchical Attention Block}
        \label{fig:block_arch}
    \end{figure*}

    The network is a fully convolutional network with an encoder-decoder architecture similar to the U-Net. Its novelty lies in the architecture of its building blocks and attention mechanisms designed for the particularities of neuronal segmentation. In order to make a straightforward comparison of our architectural improvements, with two critical components: Multi-Scale Dense Blocks (MSDBs) and a Hierarchical Attention (HA) module, as shown in Figures \ref{fig:block_arch}(A) and \ref{fig:block_arch}(B), respectively. 
    
    % \textbf{references for figure missing}

    \textit{Encoder with Multi-Scale Dense Blocks (MSDBs):} Each block is designed to capture information across different scales in parallel, allowing it to perceive local details, mid-range contours, and global structures simultaneously (as shown in Figure \ref{fig:block_arch}A). This multi-scale capability is critical, as it allows the framework to effectively detect diverse features from small, compact structures to large, extended cell structures, all within a single model.
    % Every block in the encoder contains an MSDB. MSDB consists of three parallel dilated convolution branches with dilation rates of and a 1×1 convolutional layer for feature fusion. Each parallel branch is to capture features at different scales: the d = 1 branch captures local edge details, the d = 2 branch captures mid-range contours, and the d = 4 branch captures global structural as shown in figure \ref{fig:block_arch}A. Such a multi-scale architecture is critical for detecting structures, which can be small and compact, and cell structures, which can be extended, within the same framework. 
    % \textbf{please stop writing implementation detail}

    \textit{Bottleneck with Hierarchical Attention (HA):}
    The bottleneck block, which handles the most abstract, low-resolution features, uses a Hierarchical Attention module (see Figure \ref{fig:block_arch}B). This module works in sequence, first using channel attention to decide what features are important, and then using spatial attention to determine where they are important. The resulting attention scores are applied to the input features, allowing the model to focus on the most semantically rich and spatially coherent regions. This selective focus is critical for extracting faint structures from background noise.
    % The bottleneck block that handles the most semantically rich low-resolution features consist of a Hierarchical Attention module (as shown in figure \ref{fig:block_arch}B). It is composed of two sub-modules in sequence: one channel attention module (realized as global average pooling and a sigmoid-activated dense layer) and one spatial attention module (implemented with a 7 × 7 convolution). The channel attention speculates which features are important (“what”), meanwhile, the spatial attention reasons where they are important (“where”). The obtained attention scores are then multiplied by the input feature tensor, respectively, such that the model can focus on the utmost semantically salient and spatially coherent regions, the most important for extracting the faint structures from the background noise. \textbf{please stop writing implementation detail}
    
    \textit{Decoder:} The decoder is symmetric to the encoder, progressively increasing the feature resolution at each stage. As it upsamples, it combines its features with the corresponding high-resolution features from the encoder. These encoder features are first passed through the Hierarchical Attention (HA) module to provide the decoder with attention-modulated, high-resolution guidance. The resulting merged features are then processed by the Multi-Scale Dense Block (MSDB). This final step is designed to improve region boundaries and reconnect spatial discontinuities, which is crucial for handling disjointed structures.
    % The decoder is symmetric to the encoder, and it increases the resolution of the feature map by UpSampling2D layers. The upsampled features are concatenated with the corresponding high-resolution features from the encoder Path at each level. These encoder attributes are then put through the HA module to make sure the decoder obtains attention-modulated high-resolution guidance. The merged features are next processed by the MSDB so as to improve the region boundaries, and in order to reconnect the spatial discontinuities, such as that occurred in the disjointed structures. \textbf{please stop writing implementation detail}
    
    \subsection{Loss Computation (Optimized Hybrid Morphological Loss)}
    The functions used for loss computations are a hybrid composition of four biologically motivated objectives to reflect the active process of calculating these objectives during training.
The final loss is computed as:
\begin{equation}
    \mathcal{L}_{\text{Hybrid}} = \mathcal{W}_{\text{1}}\mathcal{L}_{\text{Tversky}} + \mathcal{W}_{\text{2}}\mathcal{L}_{\text{Boundary}} + \mathcal{W}_{\text{3}}\mathcal{L}_{\text{Focal}} + \mathcal{W}_{\text{4}}\mathcal{L}_{\text{clDice}}
\end{equation}

where $\mathcal{W}_{\text{1}}$, $\mathcal{W}_{\text{2}}$, $\mathcal{W}_{\text{3}}$,$\mathcal{W}_{\text{4}}$ are empirically determined weights (0.4, 0.2, 0.3, and 0.1, respectively) that balance the contributions of each component.\\

\begin{enumerate}
    \item Tversky Loss ($\mathcal{L}_{\text{Tversky}}$): This loss addresses severe class imbalance by providing tunable control over false positives (FP) and false negatives (FN). It is a generalization of the Dice coefficient.
\begin{equation}
\mathcal{T}_{\text{Tversky}} =
\frac{
    \sum_{i=1}^{N} p_{i,1} g_{i,1} + \epsilon
}{
    \begin{aligned}
    & \sum_{i=1}^{N} p_{i,1} g_{i,1} 
    + \alpha \sum_{i=1}^{N} (1 - p_{i,0}) g_{i,1} \\
    & + \beta \sum_{i=1}^{N} p_{i,1} (1 - g_{i,0})
    \end{aligned}
}
\end{equation}

   \begin{equation}
       \mathcal{L}_{\text{Tversky}} = 1 - \mathcal{T}_{\text{Tversky}}
   \end{equation}
    where $p_{i,1}$ and $g_{i,1}$ are the predicted probability and ground truth for the foreground class at pixel i, respectively. The terms $p_{i,0}$ $g_{i,1}$ and $p_{i,1}$ $g_{i,0}$ represent the FPs and FNs, respectively. α and β are hyperparameters that control the penalty for FPs and FNs, and $\in $ is a small constant for numerical stability. For neuronal segmentation, failing to detect a faint cells (an FN) is often more detrimental than misclassifying a background pixel (an FP). By setting α=0.3 and β=0.7, we place a higher penalty on FNs, thereby increasing the model's recall and sensitivity to sparse or weakly labeled structures.\\

   \item Contour-Weighted Boundary Loss ($\mathcal{L}_{\text{Boundary}}$): This loss term provides an explicit supervisory signal for the learning of sharp and precise object boundaries. It is a coarse loss that consists of a weighted BCE and a weighted Dice loss. To begin with, we generate a contour weight map w, in which the pixels on the ground truth mask boundary are given a higher weight than that of the non-boundary pixels. The loss is then:
   \begin{equation}
       \mathcal{L}_{\text{Boundary}} = \mathcal{L}_{\text{wBCE}} + \mathcal{L}_{\text{wDice}}
   \end{equation}
    where, 
    \begin{equation}
\mathcal{L}_{\text{wBCE}} = -\frac{1}{N}\sum_{i=1}^{N} w_i [g_i \log(p_i) + (1-g_i) \log(1-p_i)]        
    \end{equation}

   In this equation:
   \begin{itemize}
       \item N is the total number of pixels.
       \item $w_{i}$ is the weight assigned to pixel $i$.
       \item $g_{i}$ is the ground truth label for pixel $i$.
       \item $p_{i}$ is the predicted probability for pixel $i$.
   \end{itemize}
   \begin{equation}
       \mathcal{L}_{\text{wDice}} = 1 - \frac{2\sum_{i=1}^{N} w_i p_i g_i + \epsilon}{\sum_{i=1}^{N} w_i p_i^2 + \sum_{i=1}^{N} w_i g_i^2 + \epsilon}
   \end{equation}
   
   In this equation:
   \begin{itemize}
       \item $w_{i}$, $p_{i}$, and $g_{i}$ are the weight, prediction, and ground truth for pixel $i$, respectively.
       \item $\epsilon$ is a small constant added for numerical stability to prevent division by zero.
   \end{itemize}
     Region-based losses, such as Dice or Tversky, can sometimes be minimized even with slightly incorrect boundaries. This explicit boundary-focused term counteracts that tendency by penalizing any deviation from the ground-truth contour, which is essential for separating adjacent cells in dense clusters.\\

    \item Focal Loss ($\mathcal{L}_{\text{Focal}}$): This loss further tackles class imbalance by focusing the training process on "hard" examples that are difficult to classify.
    \begin{equation}
        \mathcal{L}_{\text{Focal}} = -\frac{1}{N}\sum_{i=1}^{N} \alpha_t (1 - p_{t_i})^{\gamma} \log(p_{t_i})
    \end{equation}
   where $p_{t}$ is the model's estimated probability for the ground-truth class. The focusing parameter γ reduces the loss contribution from easy examples (where $p_{t}$ is high), while the balancing parameter $α_{t}$ can be used to weight classes. We initiate with γ=3.0 and α=0.8 in our system. This substantially de-weights the loss of the predominant easy-to-identify background pixels in the input prediction, so that the model needs to invest its learning capacity to learn from a smaller set of ambiguous boundary pixels and faint foreground pixels, which are critical for accurate segmentation.\\

     \item Centerline Dice Loss ($\mathcal{L}_{\text{clDice}}$): This loss is unique in its ability to enforce topological correctness, which is vital for preserving the connectivity of elongated neuronal processes. The clDice loss is based on two new metrics, Topology Precision (Tprec) and Topology Recall (Trecall):
     \begin{equation}
         T_{\text{prec}} = \frac{\left| \text{skel}(G) \cap P \right|}{\left| \text{skel}(G) \right|}  
     \end{equation}
     \begin{equation}
         T_{\text{recall}} = \frac{\left| \text{skel}(P) \cap G \right|}{\left| \text{skel}(P) \right|}
     \end{equation}

     where G is the ground truth volume, P is the predicted probability map, and skel( ) is a soft-skeletonization operator that produces a map of the structure's centerline. The clDice loss is then:
     \begin{equation}         \mathcal{L}_{\text{clDice}} = 1 - 2 \frac{T_{\text{prec}} \times T_{\text{recall}}}{T_{\text{prec}} + T_{\text{recall}}}
     \end{equation}

     This component directly penalizes breaks or gaps in tubular structures. It ensures that long, thin cell structures are segmented as continuous, connected objects rather than fragmented segments, a property that is completely ignored by all other loss functions but is critical for downstream connectivity analysis.\\
\end{enumerate}

% \subsection{Training Configuration and Optimization Strategy}
% Training is performed using the AdamW optimizer with a learning rate of 3e-4 and weight decay of 1e-4. Gradient clipping is applied with a norm of 1.0 to stabilize updates and avoid exploding gradients during early training. Training is monitored via validation loss, with early stopping (patience = 15) and ReduceLROnPlateau (patience = 5, factor = 0.5) callbacks. The batch size is set to 16 due to the high spatial resolution and model depth, but regularization via SpatialDropout2D and GroupNorm compensates for potential overfitting. Model checkpoints are saved using validation loss as the selection criterion.

\subsection{Phase-III: Inference and Refinement}

    \textbf{Postprocessing and Morphological Refinement} : 
    The model initially produces a probability map via a sigmoid activation function, which is subsequently binarized using a threshold of 0.5. To enhance spatial coherence, a morphological closing operation is applied using a 3×3 elliptical kernel. This step effectively fills small holes and bridges minor gaps within the segmented objects. Finally, connected component analysis is performed, and any objects with an area smaller than 15 pixels are removed. This area-based filtering effectively eliminates small noise artifacts and spurious false positives, ensuring that the final output preserves morphologically plausible structures.

    \textbf{Test-Time Augmentation (TTA)}:  To enhance prediction stability in the presence of input variability, we applied a Test Time Augmentation (TTA) routine in which each test image is augmented n=5 times (flips, brightness shifts), provided through the network, and the mask average is computed \cite{kimura2021understanding}. Such ensemble approach can help to alleviate variance in predictions and brittleness to variation in intensity or orientation, a typical problem faced in fluorescence microscopy due to experimental variance.

\section{Experiments}\label{s:expt}

\subsection{Dataset} \label{s:data}

\begin{figure}[H]
    \centering
    \includegraphics[width=0.8\linewidth]{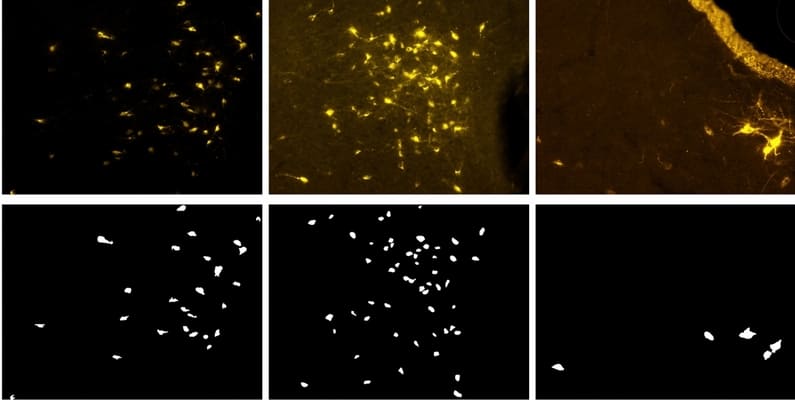}
    \caption{Few fluorescent microscopy images of brain slices with corresponding ground-truth masks.}
    \label{fig:dataset-cover}
\end{figure}

A key shift in our study was the transition from "natural images," which are typical of common benchmark datasets, to the specialized domain of biological imaging, we initially employed our research study on a dataset known as the Fluorescent Neuronal Cells dataset, containing 283 high-resolution (size 1600×1200px) microscopy images of brain slices of mice under a controlled experiment setup as shown in Figure \ref{fig:dataset-cover}.

The following dataset has been used in the work reported by Morelli et al. \cite{Morelli2021AutomatingCC} for fluorescent tracing studies of neuronal connection and structure. To specifically label neurons connected to a given anatomical region, a monosynaptic retrograde tracer, Cholera Toxin subunit B (CTb), was injected into various brain regions. The CTb was conjugated to a yellow-to-orange light-emitting fluorochrome, which was suitable for optical detection. In the images, stained neurons can be observed as bright yellowish formations in varying sizes and intensities, which are against a mainly dark background. These visual patterns allow for the discrimination of neurons from other irrelevant areas. Ground truth annotations for this dataset were developed through a hybrid-labeling method that involves manual adjustment along with semi-automated software segmentation. As mentioned in \cite{Morelli2021AutomatingCC}, the end annotations in stored as binary masks, where the tended neurons are with white pixels (value 255) and background regions are labeled as black (value 0). Such high-quality annotation guarantees accurate pixel-wise location of neuronal cells, which is critical for model training and evaluating model performance for semantic segmentation tasks. 

A summary table with average measures of main cell characteristics is reported below:

\begin{table}[h]
    \centering
    \caption{Summary of Average Measures of Main Cell Characteristics}
    \begin{tabular}{c|c}
        \toprule
        Characteristic & Value \\
        \midrule
        Area & 1206.43 \\
        Minor Axis & 29.39 \\
        Major Axis & 50.43 \\
        Equivalent Diameter & 36.50 \\
        Maximum Feret Diameter & 55.34 \\
        Mean Diameter & 47.42 \\
        \bottomrule
    \end{tabular}
    \label{tab:cell_characteristics}
\end{table}

\subsection{Comparison with Existing Research Works} \label{s:base} 
%\par\subsubsection{Comparison based on approached models\textbf{again terminology}}
% \textbf{use cite option, if you do not know how to do this meet me once today, will show you} \textbf{writing is fine for both of these}
We compare our proposed model with existing research works which we discuss next: 

% \textbf{figure 5 is not yet chnged}
\begin{enumerate}
    \item U-Net \cite{ronneberger2015u}: Traditional encoder-decoder architecture with symmetric skip connections across its layers. Even if it is effective and widely used yet its representational ability in segmenting finely detailed and complex neural structures is low.
    
    \item U-Net++ \cite{zhou2018unet++}: Extension of the U-Net model by using a combination of stacked and dense skip connections, which improves feature fusion across scales. This improvement increases the network’s capability to detect edges and the structural consistency in biomedical images.

    \item Attention U-Net \cite{Oktay2018AttentionUL}: Applies attention gates on U-Net for the salient regions of the network. It is useful to improve the performance on the sparse or local targets; however, with the fixed position of attention and the limited receptive field, it does not fit for the complicated overlaps of neurons.

    \item DeepLabV3+ \cite{Chen2018EncoderdecoderWA}: Architecture based on ASPP and a strong encoder-decoder structure. Although it scales well with the size of objects, it is not optimal for describing the fine-grained morphological continuity that is critical for the segmentation of cells.

    \item TransUNet \cite{Chen2021TransunetTM}: Adopts transformers to imitate the long-range dependency. Its segmentation performance is high for anatomically spread structures, although it is highly data-dependent and does not provide topology-preserving specific modules.

    \item MobileViT \cite{Mehta2021MobilevitLG}: A lightweight vision transformer model that integrates both convolutional efficiency and global context modelling. Although being compact and efficient but the spatial resolution is not good enough to resolve the fine structures of neurones in high-resolution microscopy images.

    \item UNeXt \cite{Valanarasu2022UNEXTMR}: State-of-the-art MLP-based design optimized for speed and hardware efficiency. However, it is not good at border recovery and fine structural segmentation, making it unsuitable for neuronal cell imaging, which requires connection and continuity.

    \item DTASUNet \cite{Ma2024DTASUNetAL}: Encoder-Decoder transformer based architecture designed for efficient medical image segmentation. It combines dynamic token aggregation in the encoder that can model long-range dependencies with a lightweight model. And it further applies channel-wise squeeze operations in the decoder to guide more precise feature recalibration. Although DTASUNet has considerable potential to segment the coarse organ boundaries under low-resource constraints, the restricted spatial correlation and the absence of hierarchical context integration reduces its performance for the fine-grained and high-resolution task such as neuronal structure segmentation.
    
\end{enumerate}
 % \textbf{very good baselines, this seems complete now}
\subsection{Comparison based on Loss Functions}
We compare the proposed hybrid loss formulation with several widely adopted loss functions commonly used in biomedical cell segmentation. Each function addresses a specific difficulty, such as class imbalance, border precision, and morphological structure preservation. The following baseline loss functions were analysed individually under the same training circumstances which we discuss next.
\begin{enumerate}

    \item Tversky Loss \cite{Salehi2017TverskyLF}: A generalization of the Dice loss that allows for tunable false positive and false negative control. We emphasize recall with α = 0.7 (weight of false negatives) and β= 0.3 (weight of false positives), which is important for recovering elongated cells.

    \item Focal Loss \cite{lin2017focal}: Loss developed to alleviate issues related to class imbalance by down-weighting easy examples, shifting focus of learning to hard areas. It leverages uncertain locations (e.g., dim or broken cells) for neuron segmentation.

    \item Centerline Dice Loss (clDice) \cite{Shi2024CenterlineBD}: Proposed with the purpose of segmenting thin, tubular structures, where expected masks are aligned with its centerlines aiming to increase structural continuity and reducing breakings in narrow neural processes.
    
    \item Contour-Weighted Boundary Loss \cite{Huang2024ContourweightedLF}: A boundary-influenced loss which emphasize on the pixels around the boundaries of objects. Dynamic reweighting of contour regions is particularly important for polymorphic neuronal borders, where the model's ability to localize small structural edges in class-imbalanced data is enhanced.

\end{enumerate}

\subsection{Ablation Experiment Study}\label{s:ablexp}
To measure the effectiveness of the proposed NeuroSegNet architecture, we compare the performance of different variants and study the role of individual losses with the proposed hybrid loss function.
\par
\subsubsection{Model Architecture}
To evaluate the impact of architectural enhancements in our proposed models, we conduct an ablation study comparing NeuroSegNet with IMSAHLO. Both models have a U-Net-like backbone, but they differ in their use of sophisticated modules and processes. Variants were trained with the same hybrid loss function to ensure consistency.

\begin{enumerate}
    \item Baseline (Residual-Only Network): This setting employs residual connections alone that are effective to improve the gradient flow and help stabilize the training process. It is the base model for comparisons and gives a performance reference and it doesn't not contain extra attention or multi-scale modules.

    \item NeuroSegNet (with SE Blocks): NeuroSegNet goes beyond our baseline by incorporating Squeeze-and-Excitation (SE) modules \cite{hu2018squeeze} for channel-wise feature recalibration that is guided by Swish activations to enhance non-linear invariant modeling. These improvements aid in better representation of complex neural cells to facilitate higher recall and improved structural coverage especially when it comes to detecting faint or thin cell regions.
    \item NeuroSegNet + MSDBs: This variant includes the Multi-Scale Dense blocks (MSDB) \cite{chang2019multi} to help the network learn features from various receptive fields. Incorporation of MSDs improves boundary precision and the model’s capability of handling coarse-grained and fine-grained structural variations, leading to higher F1 scores and superior segmentation consistencies.
    \item IMSAHLO : IMSAHLO denotes the full mode of our proposed method. Extending NeuroSegNet + MSDBs, we incorporate Hierarchical Attention (HA) mechanisms \cite{Ding2019HierarchicalAN} to adaptively highlight spatially informative areas in segmentation. This is particularly important in low contrast and overlapped cell areas. MSD blocks and HA achieve better accuracy, and classwise balanced (Macro F1) segmentation, leading to better overall morphological intactness of the segmented cells.
\end{enumerate}

\subsubsection{Loss Function}
    To evaluate the contribution of each component to our proposed hybrid loss function, we conducted a comprehensive ablation study using various combinations of the four loss terms: Tversky ($\mathcal{L}_{\text{Tversky}}$) \cite{Salehi2017TverskyLF}, Contour-Weighted ($\mathcal{L}_{\text{Contour}}$) \cite{Huang2024ContourweightedLF}, Focal ($\mathcal{L}_{\text{Focal}}$) \cite{lin2017focal}, and clDice ($\mathcal{L}_{\text{clDice}}$) \cite{Shi2024CenterlineBD}. All models were trained with identical architectural settings (IMSAHLO) to ensure that the performance changes were only due to variations of the loss function.

\begin{enumerate}
    \item $\mathcal{L}_{\text{Tversky}}$: Evaluates the quality of segmentation based on region-level overlap only. It tends to favour a recall, thus enhancing the sensitivity to faint or incomplete structures. However, it may cause over-segmentation around the boundary.

    \item $\mathcal{L}_{\text{Contour}}$: Applies Contour-Weighted Boundary Loss for the purpose of highlighting edges by addressing contour pixels. Its utilization helps in segmenting boundaries, but it has no global structural enforcement and therefore can miss internal structure.

    \item $\mathcal{L}_{\text{Focal}}$: Focuses on hard-to-assign and under-represented places, so can work well at the imbalance of classes. When it is applied independently, it has the disadvantage of introducing noisy perimeter and non closed structure.

    \item $\mathcal{L}_{\text{clDice}}$: Enhances centerline continuity so maintaining thin or elongated neural process architecture is possible. However, it is not strong at segmenting dense and overlapped regions.

    \item $\mathcal{L}_{\text{Tversky}}$ + $\mathcal{L}_{\text{Contour}}$: Applies region-level accuracy and boundary optimization to improve edge alignment, but without considering the issues of class imbalance and structure continuity.
    
    \item $\mathcal{L}_{\text{Tversky}}$ + $\mathcal{L}_{\text{Focal}}$: Combines the merits of regional sensitivity and hard example focus and produces a better foreground segmentation, however, without detailed edge-accuracy and structure connectivity.

    \item $\mathcal{L}_{\text{Tversky}}$ + $\mathcal{L}_{\text{Contour}}$ + $\mathcal{L}_{\text{Focal}}$: Combines region, boundary and imbalance-oriented losses to handle the most of the typical segmentation problems. But, the absence of clDice leads to scattered results on thin objects.

    \item $\mathcal{L}_{\text{Hybrid}}$ (Proposed): The proposed hybrid loss combines Tversky($\mathcal{L}_{\text{Tversky}}$), Contour-Weighted ($\mathcal{L}_{\text{Contour}}$), Focal($\mathcal{L}_{\text{Focal}}$), and clDice ($\mathcal{L}_{\text{clDice}}$) losses with weights of 0.4, 0.3, 0.2, and 0.1, respectively. This formulation balances region recall, edge precision, class imbalance handling, and structural continuity. 
\end{enumerate}

\subsection{Implementation Details}\label{s:imp}
\begin{figure}[H]
    \centering
    \includegraphics[width=0.8\linewidth]{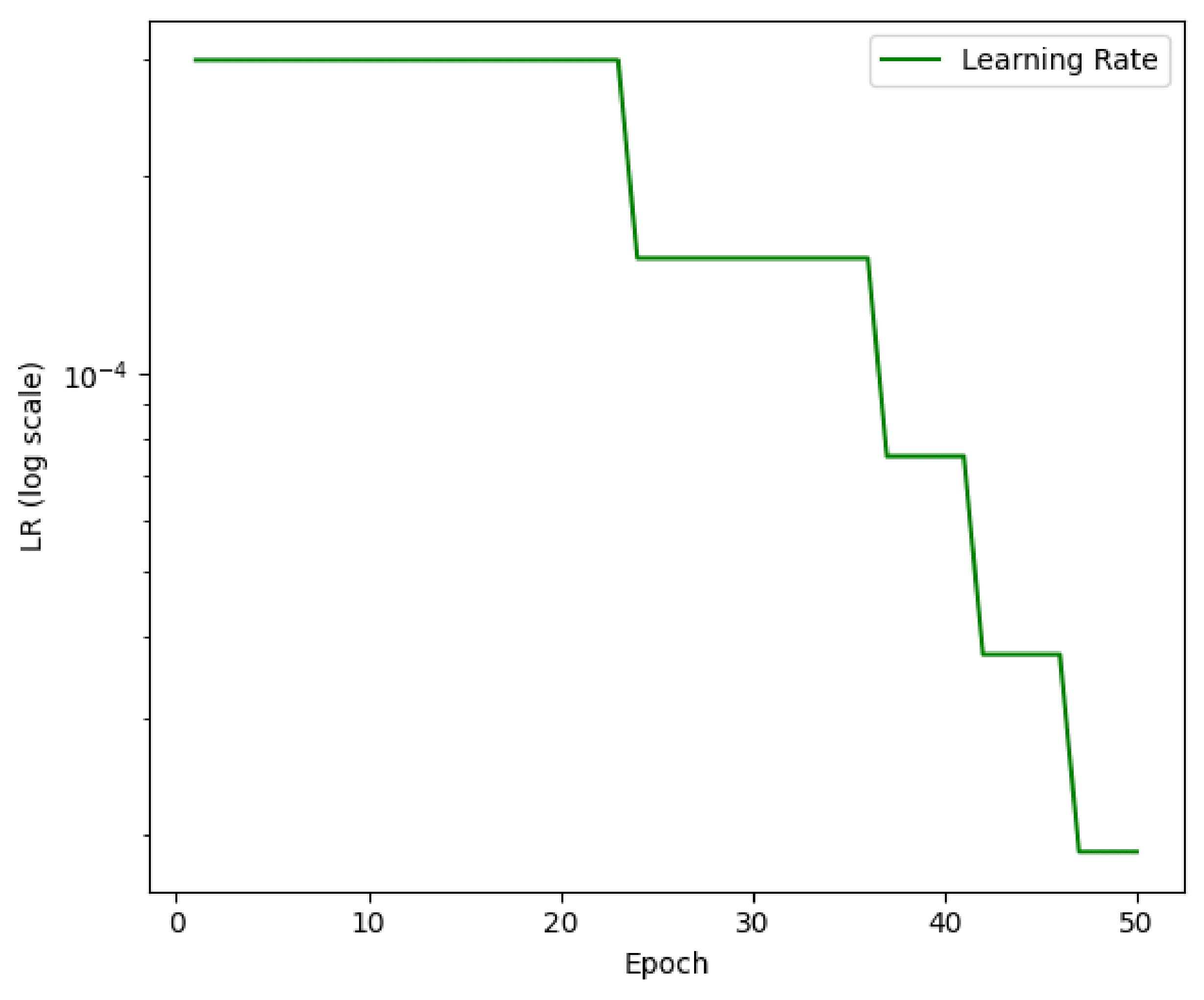}
    \caption{Learning rate schedule across training epochs.}
    \label{fig:learning_rate}
\end{figure}

% \textbf{write 2 sentences telling what are the parameters then write the values} \textbf{ this is again results or hyperparameter analysis, am sharing another paper for this how to write}.
Training was performed using the AdamW optimizer with a learning rate of $3 \times 10^{-4}$ and weight decay of $1 \times 10^{-4}$ as shown in Figure \ref{fig:learning_rate}. 
Global gradient clipping (norm = 1.0) was applied to prevent gradient explosion, especially during early training epochs. The training regimen employed several callbacks: EarlyStopping (patience = 15) to prevent overfitting, ReduceLROnPlateau with factor = 0.5 and patience = 5 to adaptively reduce the learning rate upon stagnation, and ModelCheckpoint to persist the best model based on validation loss as the selection criterion. Training was performed for 50 epochs with a batch size of 16, justified by GPU memory constraints ($<$12 GB RAM) for the input image of dimensions $512 \times 768$. However, due to efficient regularization(GroupNorm, SpatialDropout), the network remained stable and generalized well across epochs. We use TensorFlow libraries for implementation. This implementation helped confirm that the network successfully captured thin cell structures and preserved morphology.

\section{Results and Discussions}\label{s:res}

\begin{table*}[t]
\small
\setlength{\tabcolsep}{2pt}
\centering
\caption{Training and Validation Performance of Baseline Models}
\label{tab:baseline_models}
\begin{tabular}{|l|c|c|c|c|c|c|c|c|c|c|c|c|}
\hline
\multirow{2}{*}{\textbf{Model}} & \multicolumn{6}{c|}{\textbf{Training}} & \multicolumn{6}{c|}{\textbf{Validation}} \\
\cline{2-13}
& Accuracy & Precision & Recall & F1 Score & Macro F1 & Micro F1 & Accuracy & Precision & Recall & F1 Score & Macro F1 & Micro F1 \\
\hline
U-Net (2015)           & 0.969 & 0.652 & 0.663 & 0.658 & 0.618 & 0.647 & 0.962 & 0.637 & 0.658 & 0.647 & 0.539 & 0.597 \\
U-Net++ (2018)         & 0.978 & 0.678 & 0.684 & 0.681 & 0.683 & 0.692 & 0.970 & 0.678 & 0.659 & 0.668 & 0.654 & 0.665 \\
Attention U-Net (2018) & 0.979 & 0.705 & 0.709 & 0.707 & 0.716 & 0.721 & 0.974 & 0.710 & 0.715 & 0.712 & 0.705 & 0.713 \\
DeepLabV3+ (2018)      & 0.981 & 0.715 & 0.712 & 0.714 & 0.721 & 0.728 & 0.977 & 0.717 & 0.710 & 0.713 & 0.708 & 0.718 \\
MobileViT (2021)       & 0.974 & 0.685 & 0.678 & 0.681 & 0.693 & 0.704 & 0.968 & 0.695 & 0.683 & 0.689 & 0.678 & 0.690 \\
TransUNet (2021)       & 0.985 & 0.724 & 0.717 & 0.720 & 0.729 & 0.732 & 0.980 & 0.727 & 0.719 & 0.723 & 0.720 & 0.766 \\
UNeXt (2023)           & 0.976 & 0.725 & 0.714 & 0.719 & 0.735 & 0.726 & 0.971 & 0.737 & 0.712 & 0.724 & 0.755 & 0.759 \\
DTASUnet (2024)       & 0.984 & 0.717 & 0.728 & 0.717 & 0.759 & 0.751 & 0.979 & 0.720 & 0.735 & 0.730 & 0.784 & 0.796 \\
IMSAHLO & \textbf{0.993} & \textbf{0.781} & \textbf{0.764} & \textbf{0.772} & \textbf{0.834} & \textbf{0.838} & \textbf{0.995} & \textbf{0.814} & \textbf{0.789} & \textbf{0.801} & \textbf{0.827} & \textbf{0.833} \\
\hline
\end{tabular}
\end{table*}

 % \textbf{a. In Table II, please remove accuracy, its dilluting your results and add the last or first row for IMSAHLO from Table III. Then make bold which is maximum, it would be IMSAHLO b. Table III is fine for now. c. Table IV mention either in caption or in first column that all these losses are done on IMSAHLO}

\begin{table*}[t]
\footnotesize
\setlength{\tabcolsep}{1.5pt}
\renewcommand{\arraystretch}{0.95}
\centering
\caption{Training and Validation Performance of Proposed Models (Ablation Study)}
\label{tab:ablation_results}
\begin{tabular}{|p{4.5cm}<{\raggedright\arraybackslash}|c|c|c|c|c|c|c|c|c|c|c|c|}
\hline
\multirow{2}{*}{\textbf{Model}} & \multicolumn{6}{c|}{\textbf{Training}} & \multicolumn{6}{c|}{\textbf{Validation}} \\
\cline{2-13}
& Accuracy & Precision & Recall & F1 Score & Macro F1 & Micro F1 & Accuracy & Precision & Recall & F1 Score & Macro F1 & Micro F1 \\
\hline
\makecell[l]{Baseline (Only Residual Blocks)} & 0.978 & 0.678 & 0.684 & 0.681 & 0.683 & 0.692 & 0.970 & 0.678 & 0.659 & 0.668 & 0.654 & 0.665 \\
\makecell[l]{NeuroSegNet (with SE Blocks)} & 0.991 & 0.752 & \textbf{0.793} & 0.771 & 0.810 & 0.815 & 0.991 & 0.767 & \textbf{0.804} & 0.785 & 0.818 & 0.828 \\
\makecell[l]{NeuroSegNet + MSDBs} & 0.992 & 0.765 & 0.783 & 0.774 & 0.822 & 0.827 & 0.993 & 0.785 & 0.797 & 0.791 & 0.822 & 0.830 \\
\makecell[l]{IMSAHLO \\ (NeuroSegNet + MSDBs + HA)} & \textbf{0.993} & \textbf{0.781} & 0.764 & \textbf{0.772} & \textbf{0.834} & \textbf{0.838} & \textbf{0.995} & \textbf{0.814} & 0.789 & \textbf{0.801} & \textbf{0.827} & \textbf{0.833} \\
\hline
\end{tabular}
\end{table*}

\begin{table*}[t]
\centering
\caption{Ablation Study on Loss Function Variants for Neuronal Cell Segmentation}
\label{tab:loss_ablation}
\begin{tabular}{|l|c|c|c|c|}
\hline
\textbf{Loss Function Variant} & \textbf{DSC} & \textbf{Macro F1} & \textbf{Micro F1} & \textbf{IoU} \\
\hline
$\mathcal{L}_{\text{Tversky}}$                         & 0.782 & 0.760 & 0.781 & 0.703 \\
$\mathcal{L}_{\text{Contour}}$                         & 0.738 & 0.728 & 0.734 & 0.670 \\
$\mathcal{L}_{\text{Focal}}$                            & 0.716 & 0.700 & 0.715 & 0.652 \\
$\mathcal{L}_{\text{clDice}}$                           & 0.722 & 0.704 & 0.720 & 0.660 \\
$\mathcal{L}_{\text{Tversky}} + \mathcal{L}_{\text{Contour}}$               & 0.785 & 0.774 & 0.784 & 0.725 \\
$\mathcal{L}_{\text{Tversky}} + \mathcal{L}_{\text{Focal}}$                 & 0.775 & 0.765 & 0.770 & 0.712 \\
$\mathcal{L}_{\text{Tversky}} + \mathcal{L}_{\text{Contour}} + \mathcal{L}_{\text{Focal}}$ & 0.789 & 0.775 & 0.793 & 0.758 \\
$\mathcal{L}_{\text{Hybrid}}$ (Proposed)& \textbf{0.801} & \textbf{0.827} & \textbf{0.833} & \textbf{0.784} \\
\hline
\end{tabular}
\end{table*}

\subsection{Comparison with Baselines}
% \textbf{1. Is Neurosegnet++ your proposed approach? 2. ablation study and baselines should be different tables. 3. training and validation should be single table, so combine in that matter. 4. dont keep Accuracy, Loss in table}
We evaluate the performance of seven commonly used segmentation models—U-Net, U-Net++, Attention U-Net, DeepLabV3+, MobileViT, TransUNet, UNeXt, and DTASUnet—on the fluorescent neuronal cell segmentation task. These baselines were compared using six evaluation metrics: Accuracy, Precision, Recall, F1 Score, Macro F1, and Micro F1, with both training and validation results summarized in Table~\ref{tab:baseline_models}.\\
Our proposed model, IMSAHLO, outperformed all baseline models across all evaluation metrics. It achieved a validation F1 Score of 0.806, a Macro F1 of 0.827, and a Micro F1 of 0.833, surpassing the strongest baseline, DTASUnet, by 8.4\%, 5.3\%, and 4.4\%, respectively. These results substantiate the effectiveness of our proposed MSD blocks and Hierarchical Attention mechanism in modeling both coarse and fine-grained cellular features within noisy, high-resolution imaging conditions.\\
For comparison, the performance of several baseline models was evaluated. The foundational U-Net model, serving as an initial baseline, achieved a validation F1 Score of 0.647 and a Macro F1 of 0.539, indicating limitations in segmenting complex neuronal morphologies. U-Net++, which utilizes nested skip connections, yielded a marginal performance increase with validation F1 and Macro F1 scores of 0.668 and 0.654, respectively, but did not significantly improve on structural and class-sensitive metrics.\\
Architectures incorporating attention mechanisms, such as Attention U-Net and DeepLabV3+, produced more balanced results. Attention U-Net exhibited a validation F1 score of 0.712, and DeepLabV3+ scored 0.713. While they surpassed their predecessors, both models still lacked the requisite sensitivity to accurately segment overlapping cell bodies.\\
Transformer-based models demonstrated superior performance over purely convolutional architectures. MobileViT achieved a validation F1 Score of 0.689 and a Macro F1 of 0.678, though its lightweight design appeared to limit its capacity for extracting high-resolution spatial data. UNeXt achieved 7.7\% and 15.4\% higher F1 Score and Macro F1 than U-Net++. TransUNet, which employs a hybrid transformer-convolutional backbone, achieved the highest F1 (0.723) and Micro F1 (0.766) scores among all the baseline models, outperforming U-Net++ by 8.2\% and 15.2\%, respectively.\\
Among all baselines, DTASUnet was established as the strongest competitor, reporting the highest validation F1 Score (0.730), Recall (0.735), and Macro F1 (0.784). Relative to U-Net++, this corresponds to a 9.2\% increase in F1 Score, an 11.5\% increase in Recall, and a 19.8\% increase in Macro F1.

\begin{figure*}[t]
    \centering
    \begin{subfigure}[t]{0.48\textwidth}
        \centering
        \includegraphics[height=5.5cm, width=\linewidth]{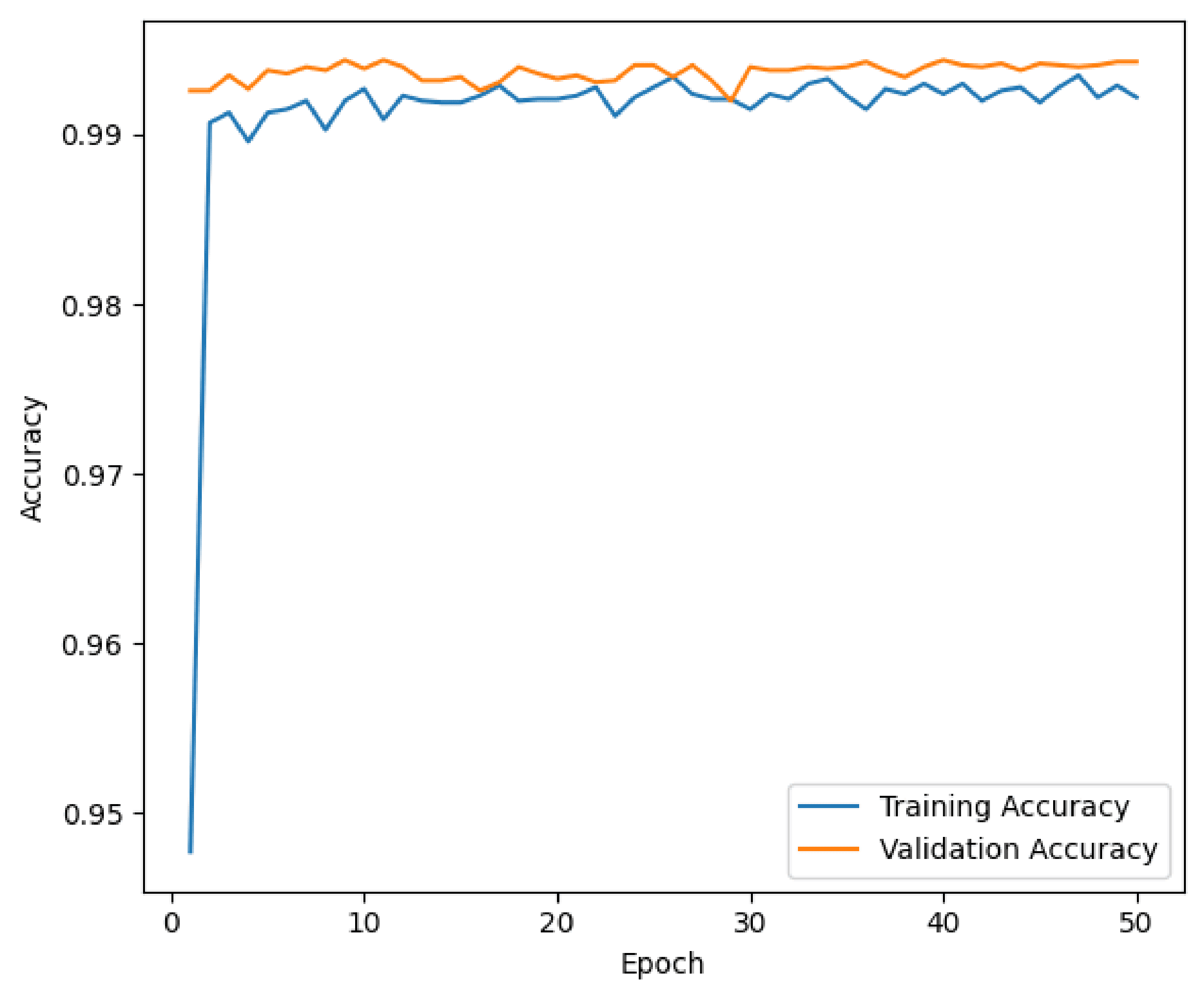}
        \caption{Accuracy over Epochs}
    \end{subfigure}
    \hfill
    \begin{subfigure}[t]{0.48\textwidth}
        \centering
        \includegraphics[height=5.5cm, width=\linewidth]{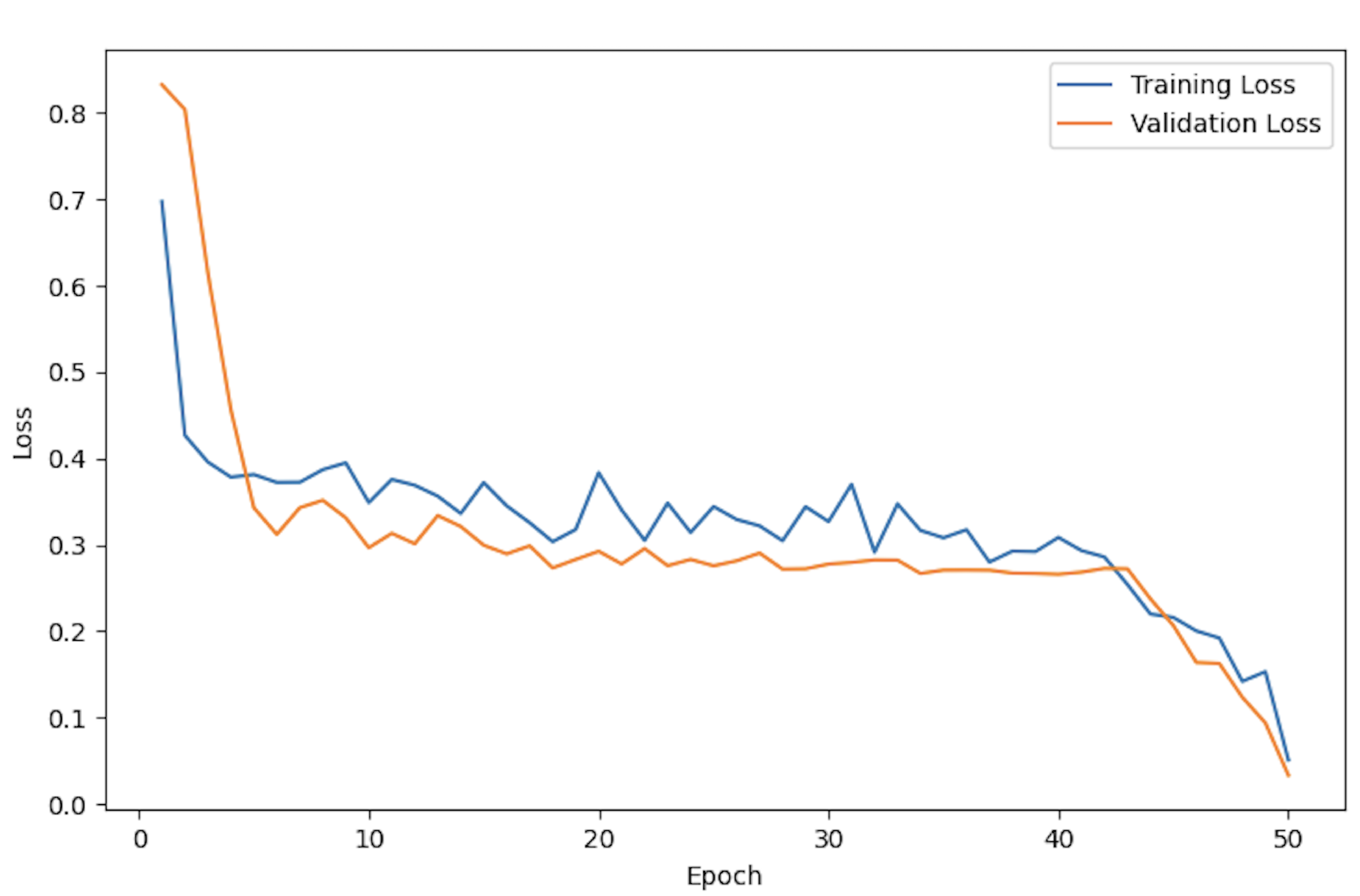}
        \caption{Loss over Epochs}
    \end{subfigure}
   %\vspace{2mm}

    \begin{subfigure}[t]{0.48\textwidth}
        \centering
        \includegraphics[height=5.5cm, width=\linewidth]{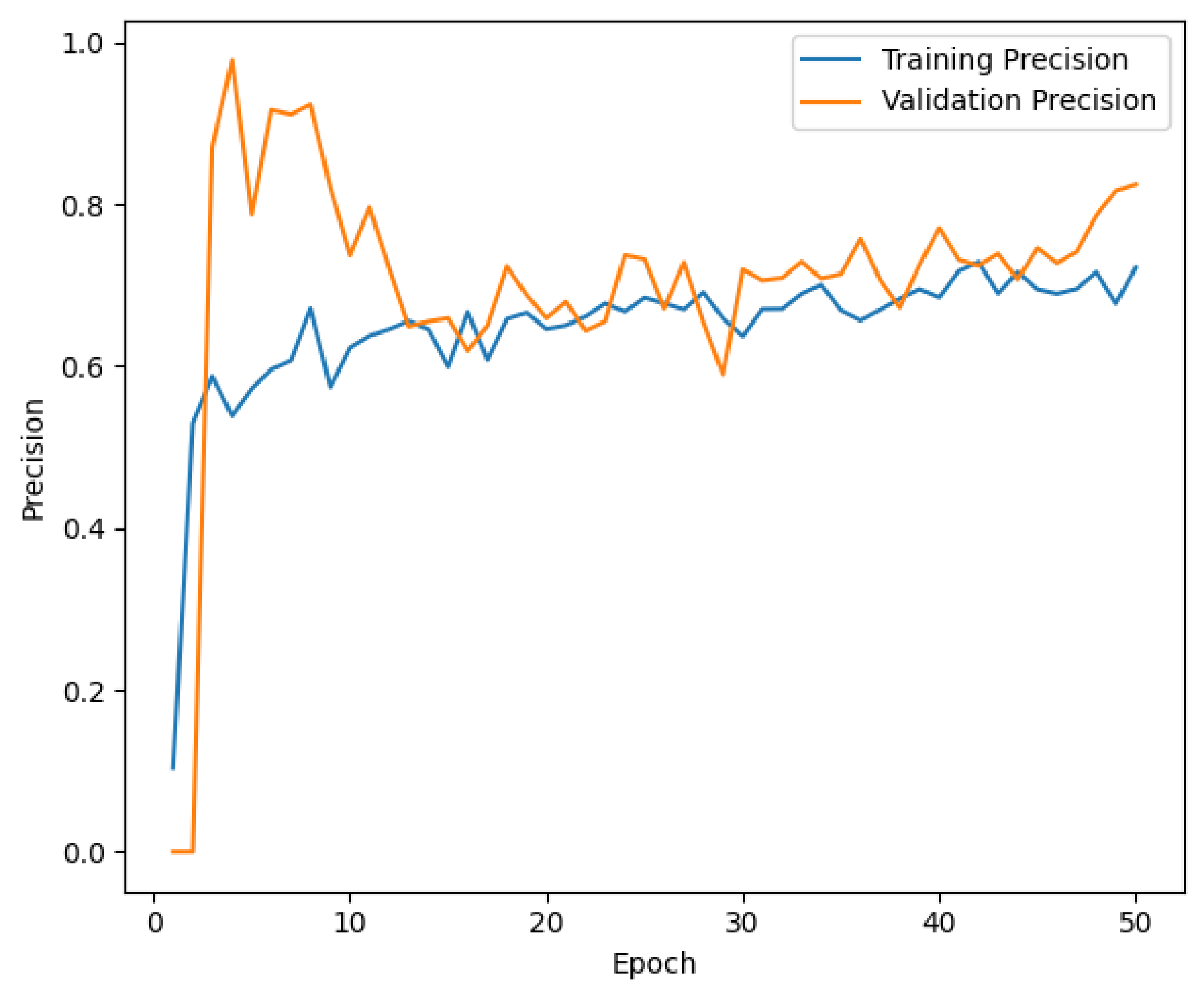}
        \caption{Precision over Epochs}
    \end{subfigure}
    \hfill
    \begin{subfigure}[t]{0.48\textwidth}
        \centering
        \includegraphics[height=5.5cm, width=\linewidth]{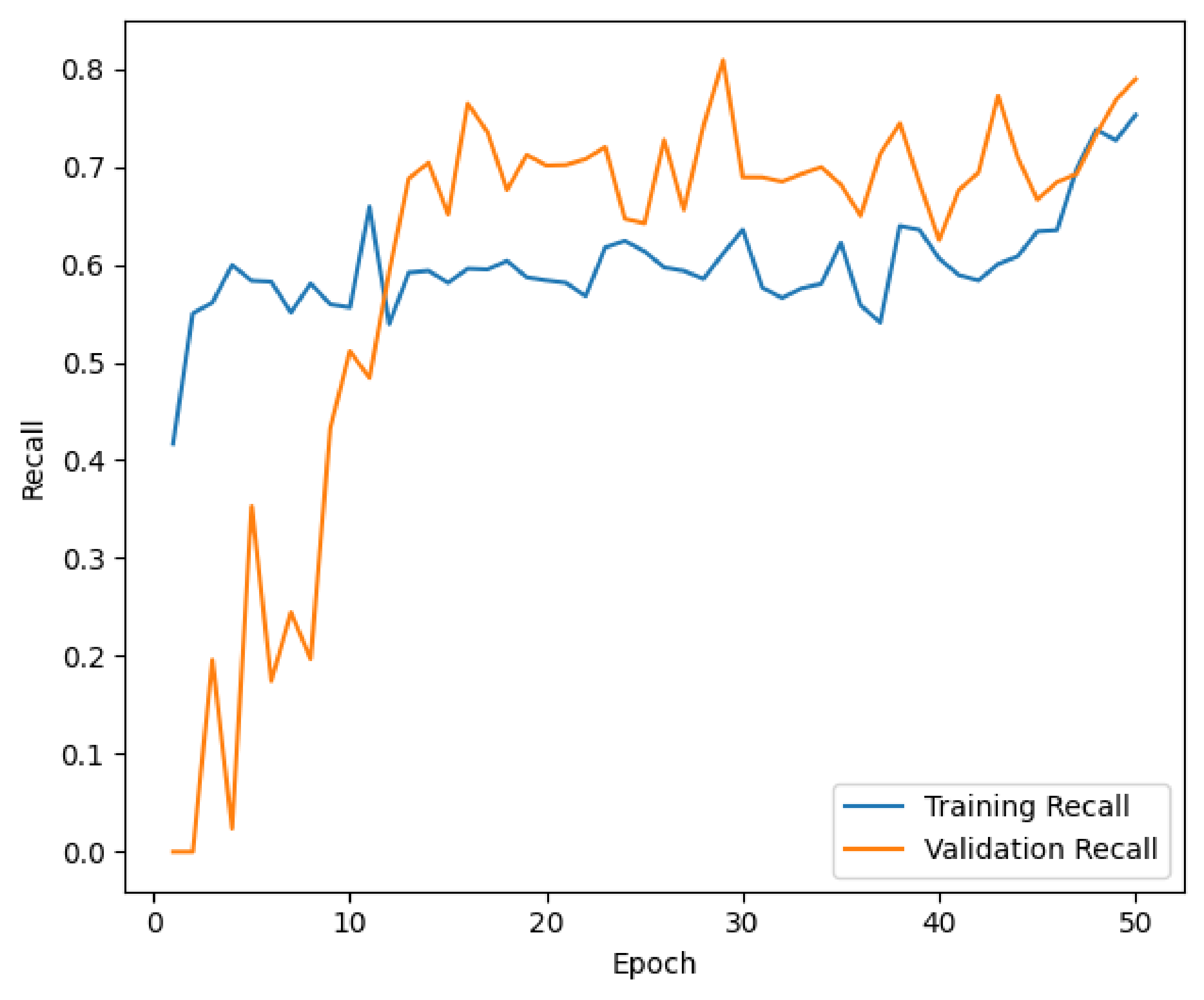}
        \caption{Recall over Epochs}
    \end{subfigure}

    \caption{Progress plots showing the evolution of Accuracy, Loss, Precision, and Recall across 50 epochs for IMSAHLO.}
    \label{fig:graphs}
\end{figure*}

\subsection{Ablation Study}
\subsubsection{Ablation Study on Model Architecture}
The architectural ablation results for NeuroSegNet and IMSAHLO are shown in Table \ref{tab:ablation_results}. This study systematically evaluates the performance impact of progressively integrating key modules: residual connections, Swish activations, Squeeze-and-Excitation (SE) blocks, MultiScaleDense (MSD) blocks, and Hierarchical Attention.\\
Our baseline model, equipped solely with residual connections, serves as the foundational benchmark. It achieved a validation F1 Score of 0.668 and a Macro F1 of 0.654, representing the initial segmentation capability before the inclusion of advanced multi-scale or attention-based mechanisms.\\
The integration of SE blocks into the baseline to form NeuroSegNet yielded substantial improvements, achieving a validation Precision of 0.767 and Recall of 0.804. This resulted in a validation F1 Score of 0.785, indicating enhanced detection of cellular structures, particularly those that are small or faint. The corresponding increase in Macro F1 to 0.818 suggests improved class-wise performance, attributable to the channel-wise feature recalibration provided by the SE modules.\\
Further augmenting the architecture with MSD blocks (NeuroSegNet + MSDBs) produced additional performance gains. This configuration achieved a validation F1 Score of 0.791, with a Precision of 0.785 and a Recall of 0.797. The improvement is attributed to the MSD blocks' ability to capture features across multiple receptive fields, leading to better boundary definition and detection of structural variations at different scales. The Macro F1 also rose to 0.822, reflecting greater consistency across classes.\\
Finally, the complete IMSAHLO architecture, which combines residual connections, MSD blocks, and Hierarchical Attention, delivered the strongest overall performance. It recorded a validation F1 Score of 0.801, Precision of 0.814, and Macro F1 of 0.827. This represents a 3.3\% absolute improvement in F1 score over the baseline and 1.6\% over NeuroSegNet. The Hierarchical Attention mechanism enhances spatial adaptivity, enabling the model to focus on salient regions, which significantly improves segmentation accuracy and reduces false positives in challenging low-contrast or overlapping cell areas.\\
Notably, while NeuroSegNet achieved the highest Recall (0.804 vs. IMSAHLO’s 0.789), suggesting superior sensitivity to faint or extended cells, this came at the cost of lower precision due to a higher rate of false positives. In summary, the ablation study demonstrates that while NeuroSegNet excels in raw sensitivity, IMSAHLO offers a more balanced and precise segmentation. Its integration of multi-scale feature aggregation and spatial attention results in superior boundary localization, a lower false positive rate, and improved overall morphological integrity.
\begin{figure*}[t]
    \centering

    % Dense Example 1
    \begin{subfigure}[t]{0.16\textwidth}
        \includegraphics[width=\linewidth, height=2.8cm]{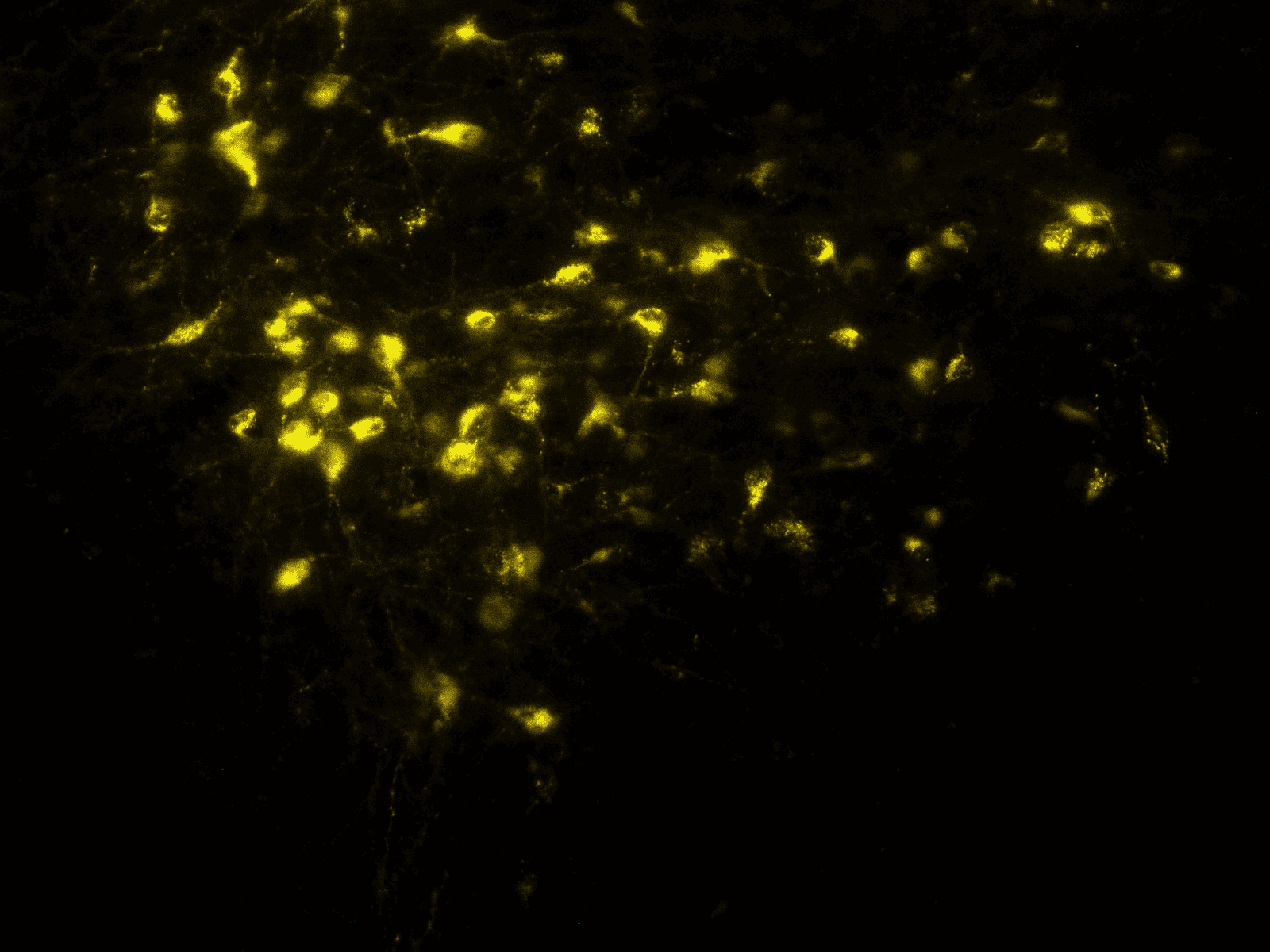}
        \caption*{Input (Dense 1)}
    \end{subfigure}
    \begin{subfigure}[t]{0.16\textwidth}
        \includegraphics[width=\linewidth, height=2.8cm]{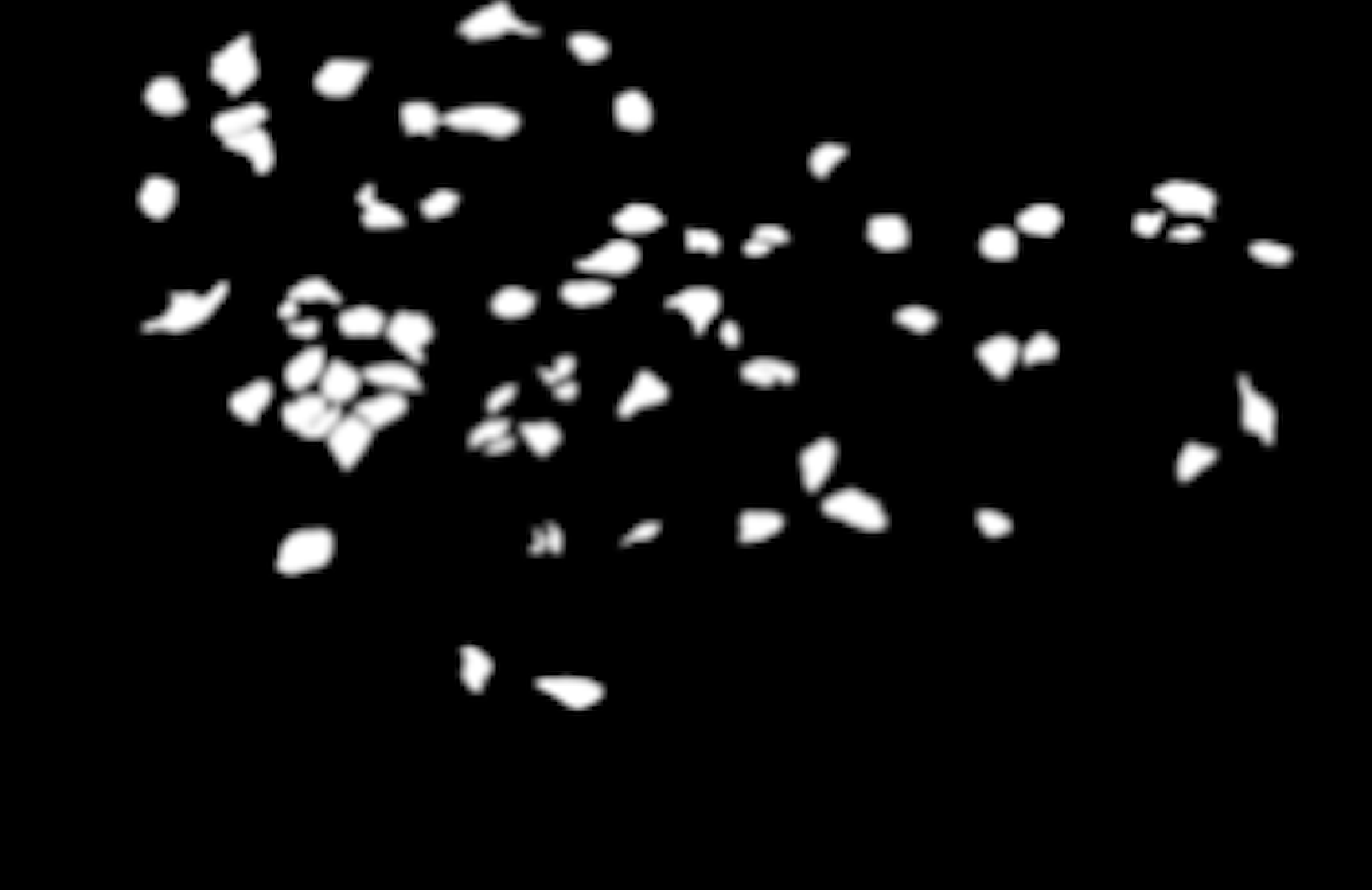}
        \caption*{GT}
    \end{subfigure}
    \begin{subfigure}[t]{0.16\textwidth}
        \includegraphics[width=\linewidth, height=2.8cm]{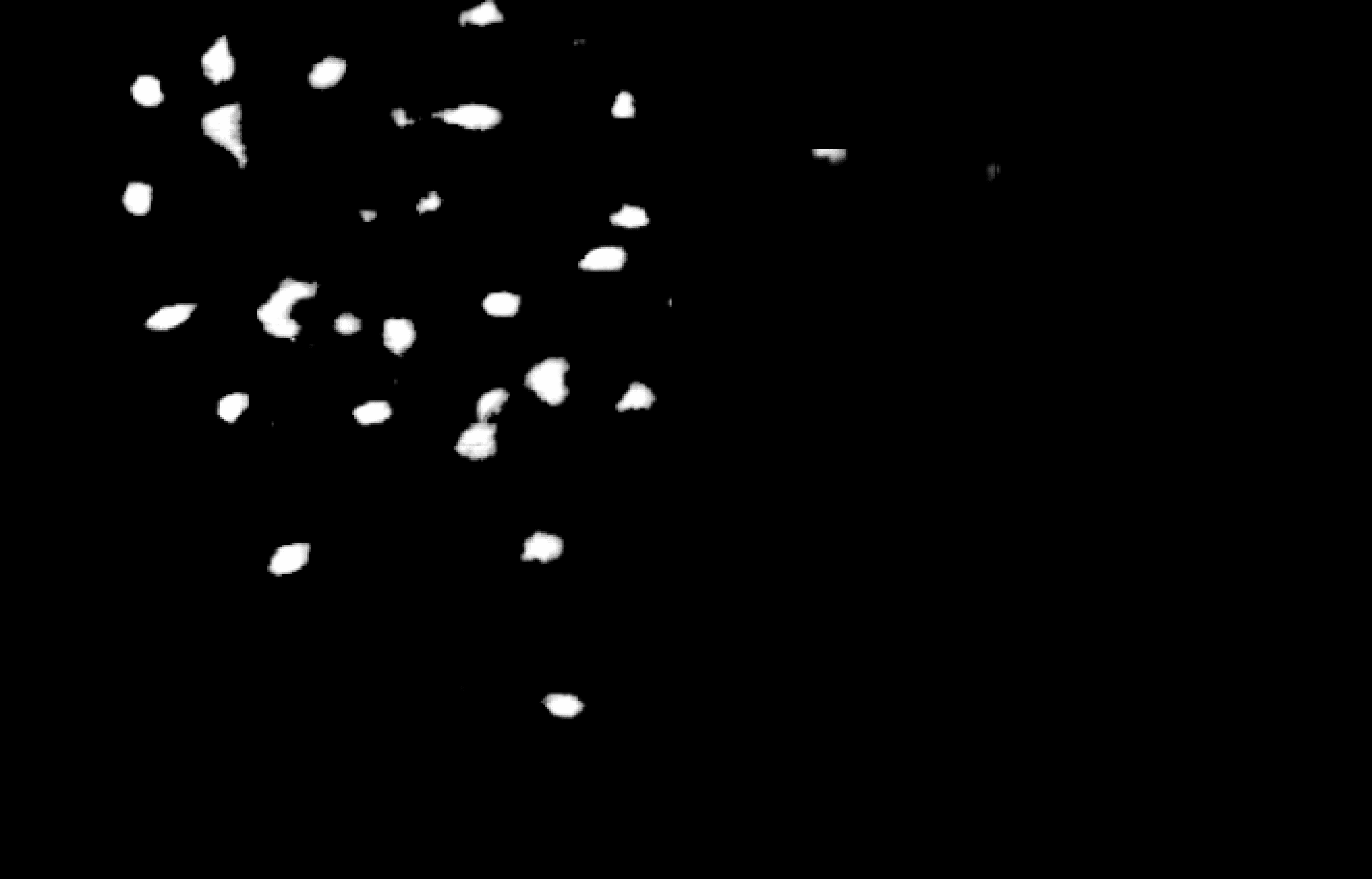}
        \caption*{TransUNet}
    \end{subfigure}
    \begin{subfigure}[t]{0.16\textwidth}
        \includegraphics[width=\linewidth, height=2.8cm]{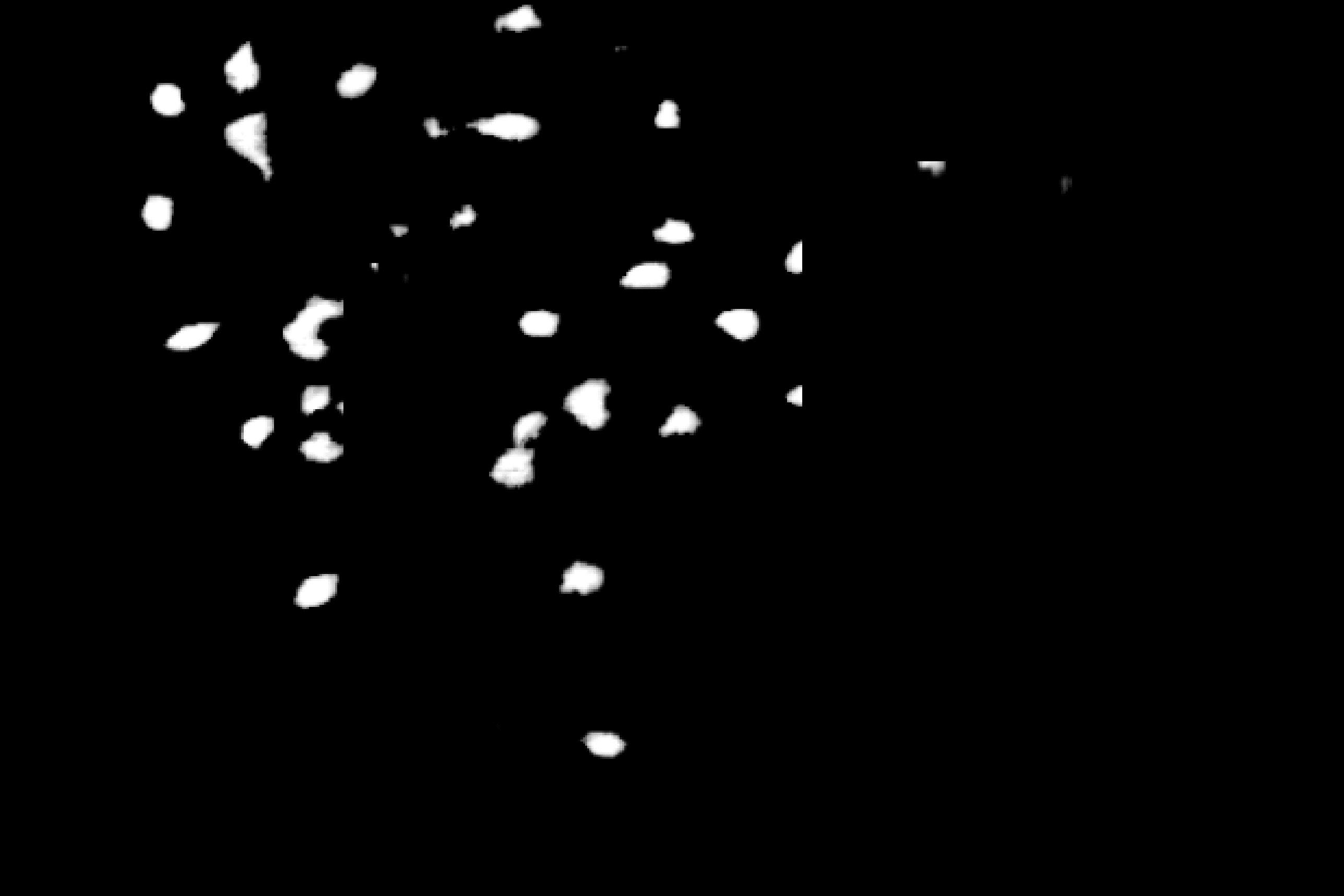}
        \caption*{DTASUNet}
    \end{subfigure}
    \begin{subfigure}[t]{0.16\textwidth}
        \includegraphics[width=\linewidth, height=2.8cm]{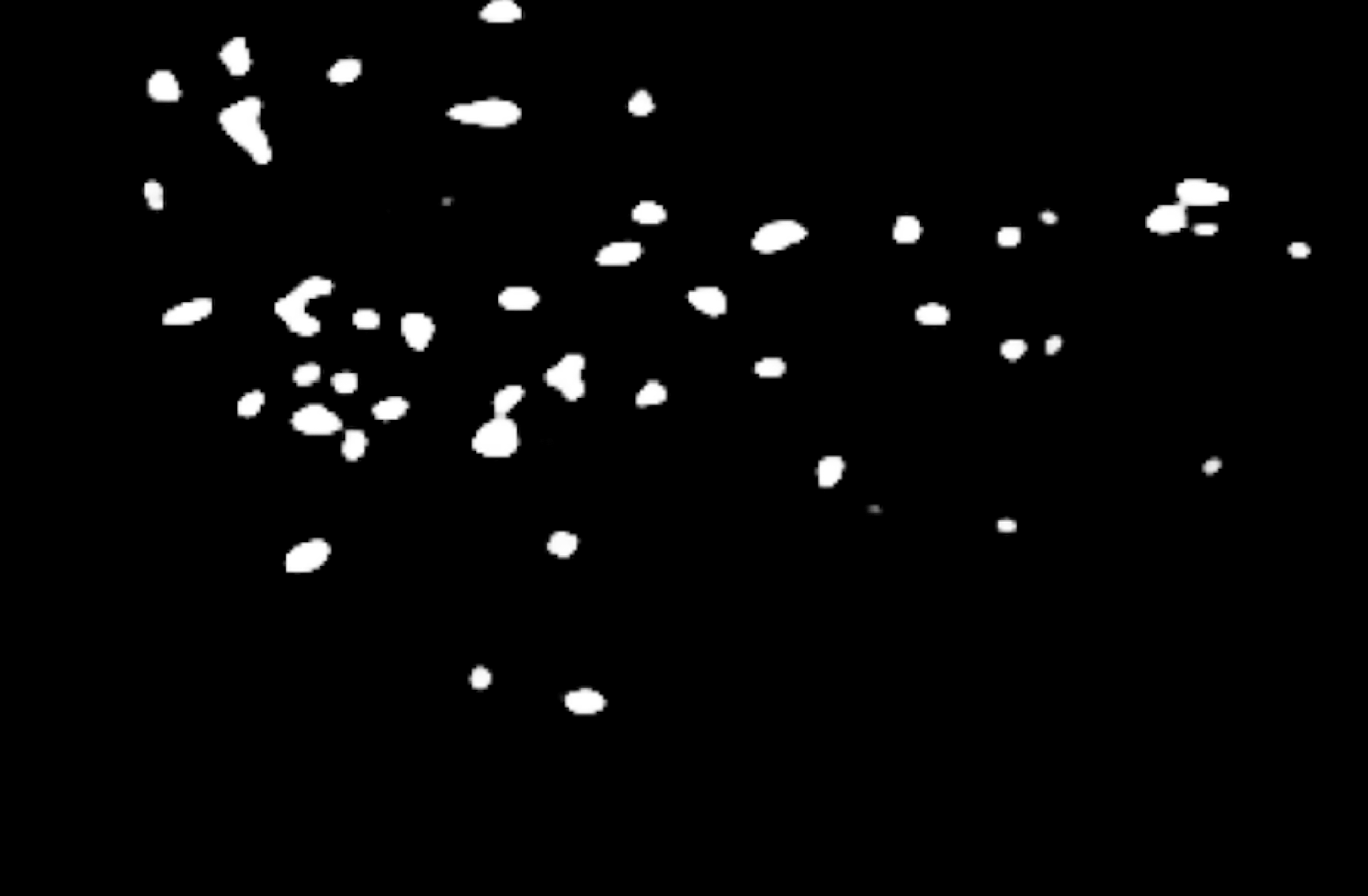}
        \caption*{NeuroSegNet}
    \end{subfigure}
    \begin{subfigure}[t]{0.16\textwidth}
        \includegraphics[width=\linewidth, height=2.8cm]{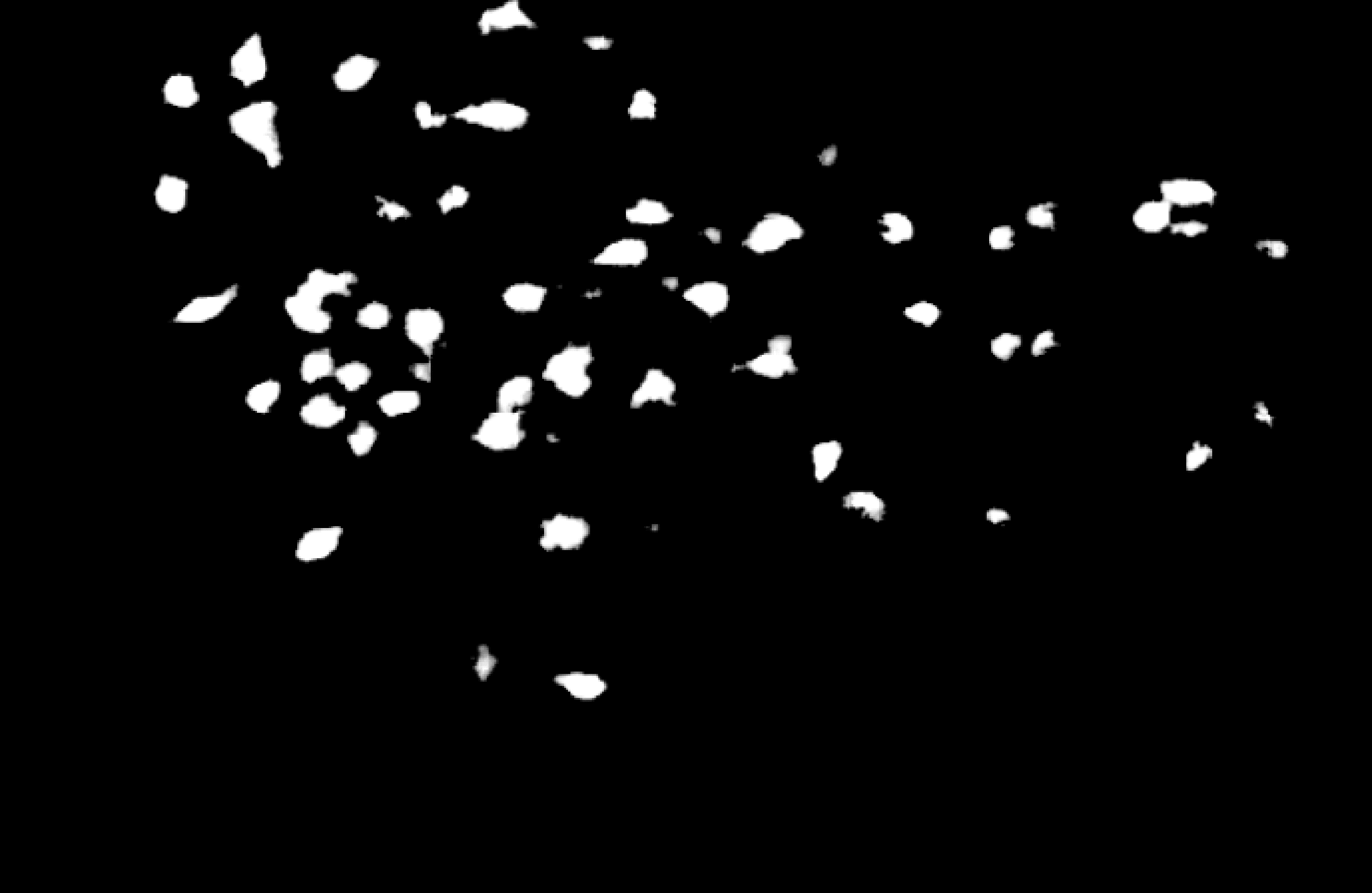}
        \caption*{IMSAHLO}
    \end{subfigure}

    % Dense Example 2
    \vspace{2mm}
    \begin{subfigure}[t]{0.16\textwidth}
        \includegraphics[width=\linewidth, height=2.8cm]{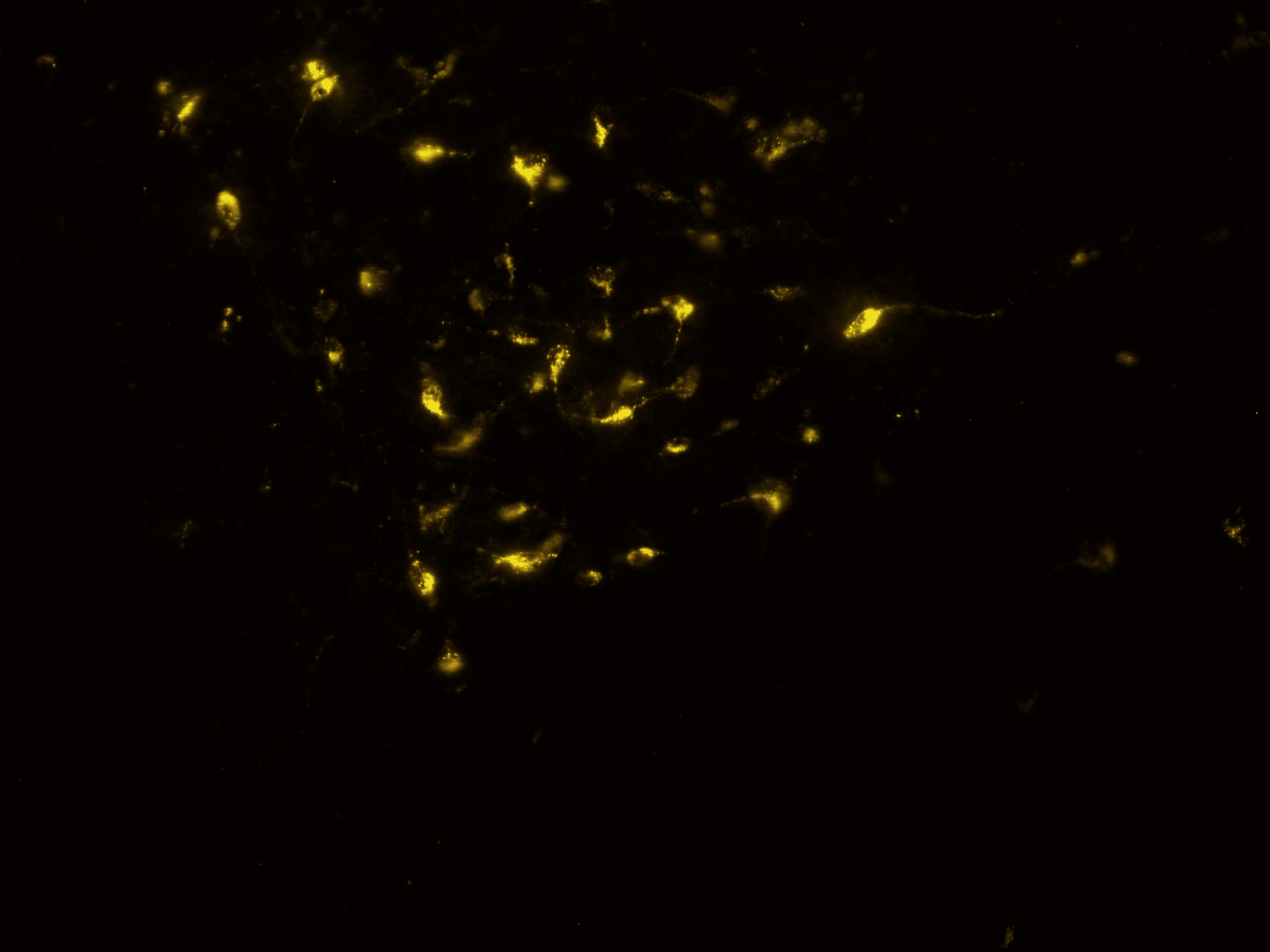}
        \caption*{Input (Dense 2)}
    \end{subfigure}
    \begin{subfigure}[t]{0.16\textwidth}
        \includegraphics[width=\linewidth, height=2.8cm]{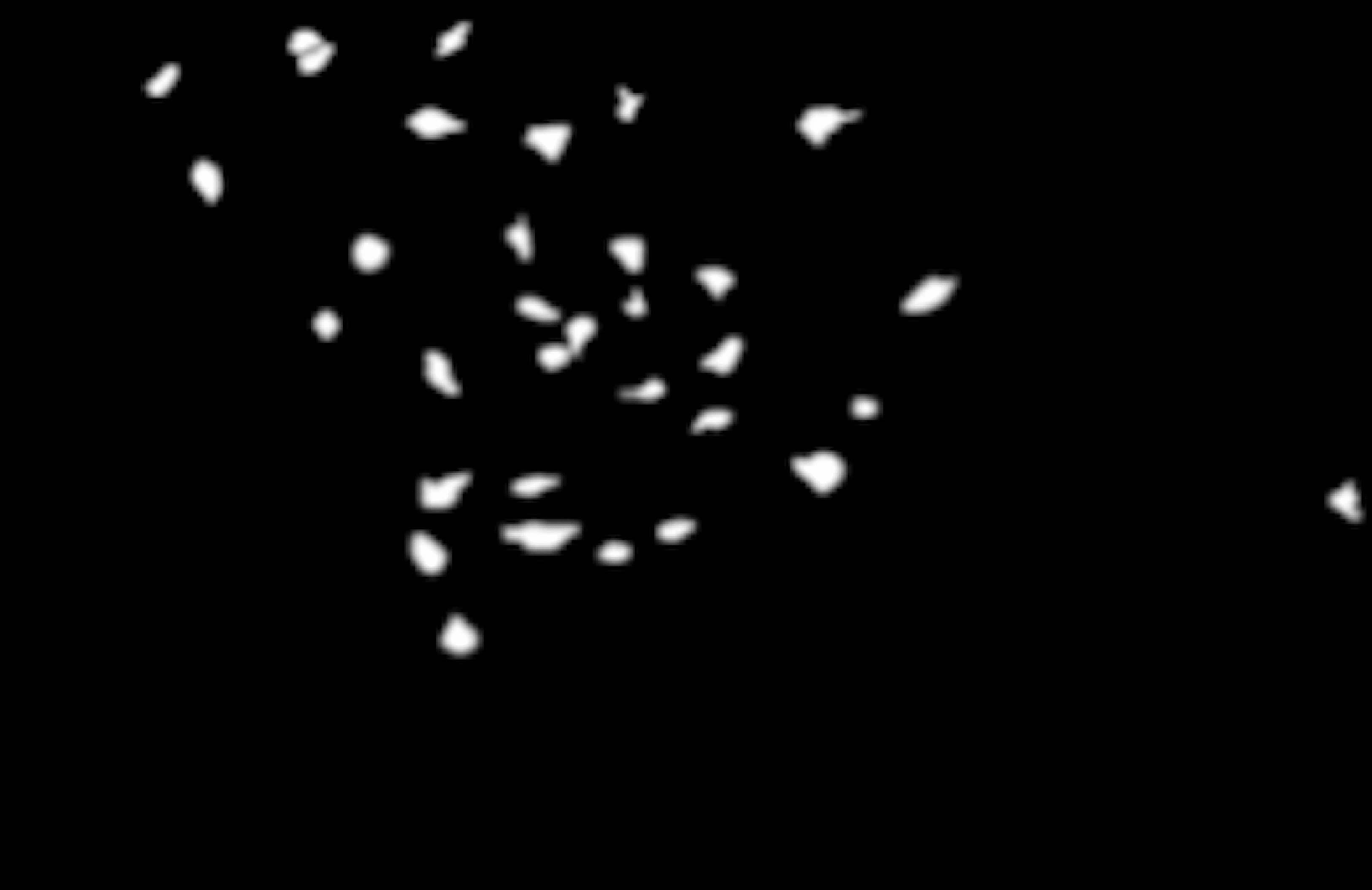}
        \caption*{GT}
    \end{subfigure}
    \begin{subfigure}[t]{0.16\textwidth}
        \includegraphics[width=\linewidth, height=2.8cm]{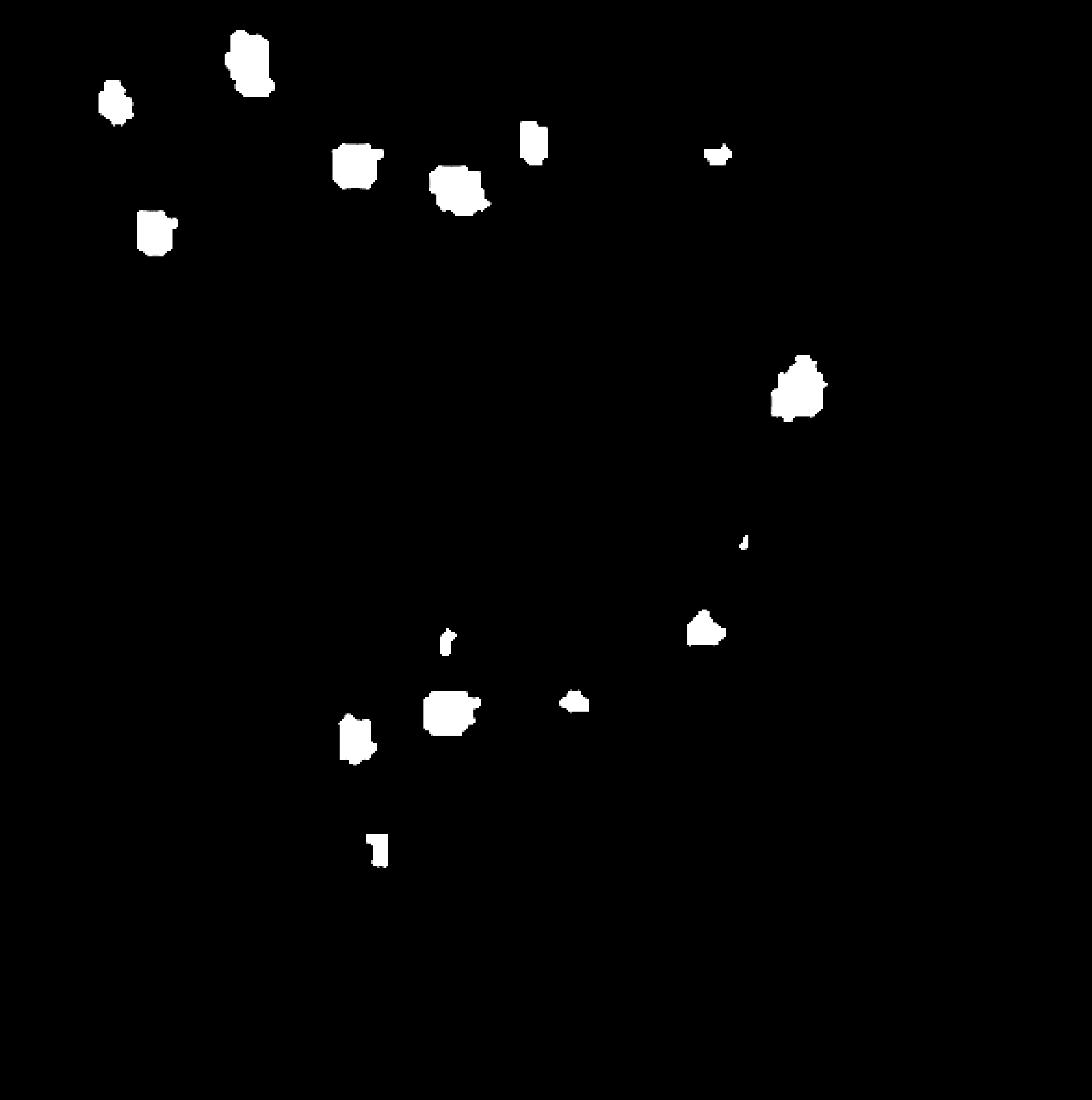}
        \caption*{TransUNet}
    \end{subfigure}
    \begin{subfigure}[t]{0.16\textwidth}
        \includegraphics[width=\linewidth, height=2.8cm]{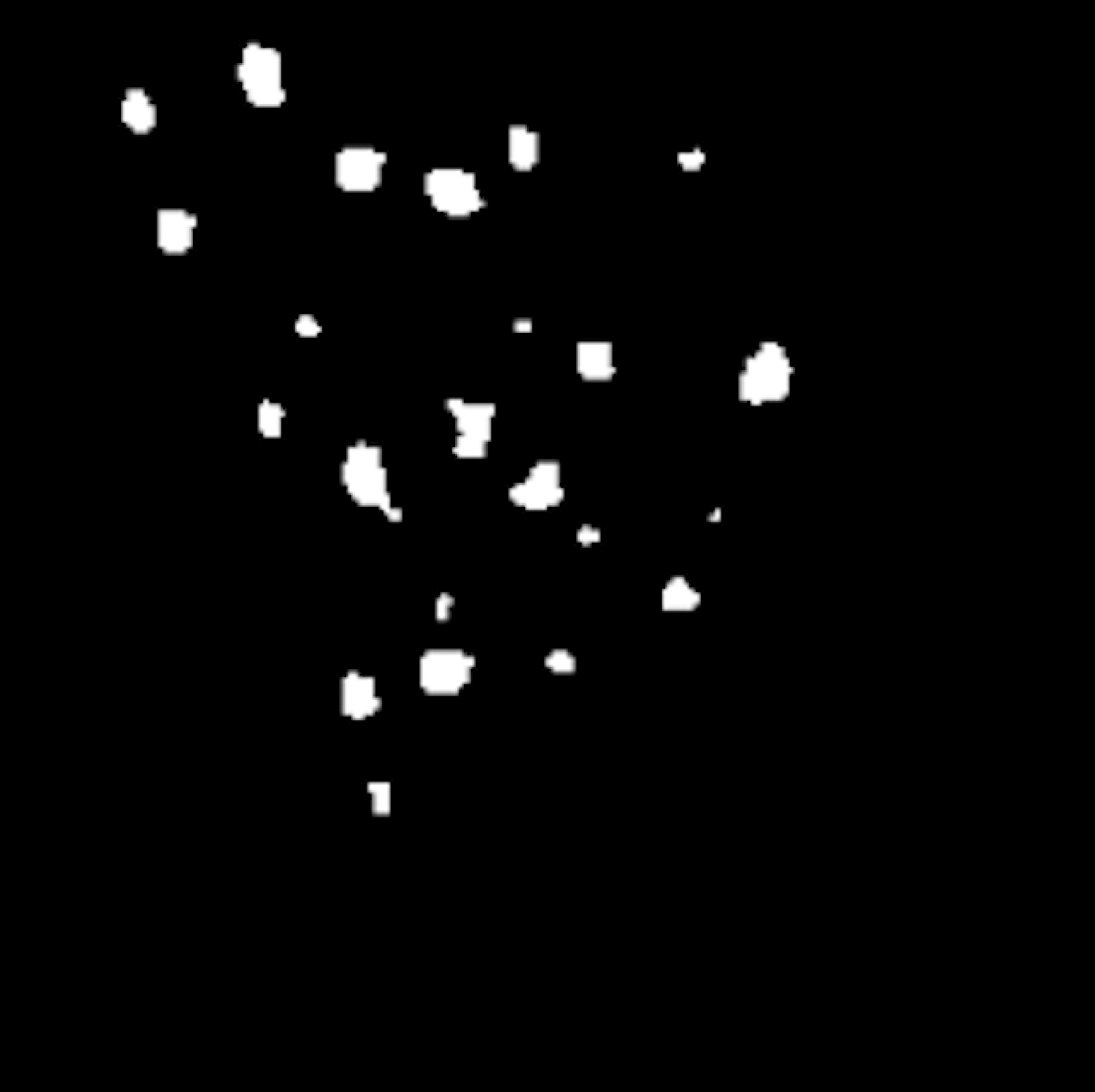}
        \caption*{DTASUNet}
    \end{subfigure}
    \begin{subfigure}[t]{0.16\textwidth}
        \includegraphics[width=\linewidth, height=2.8cm]{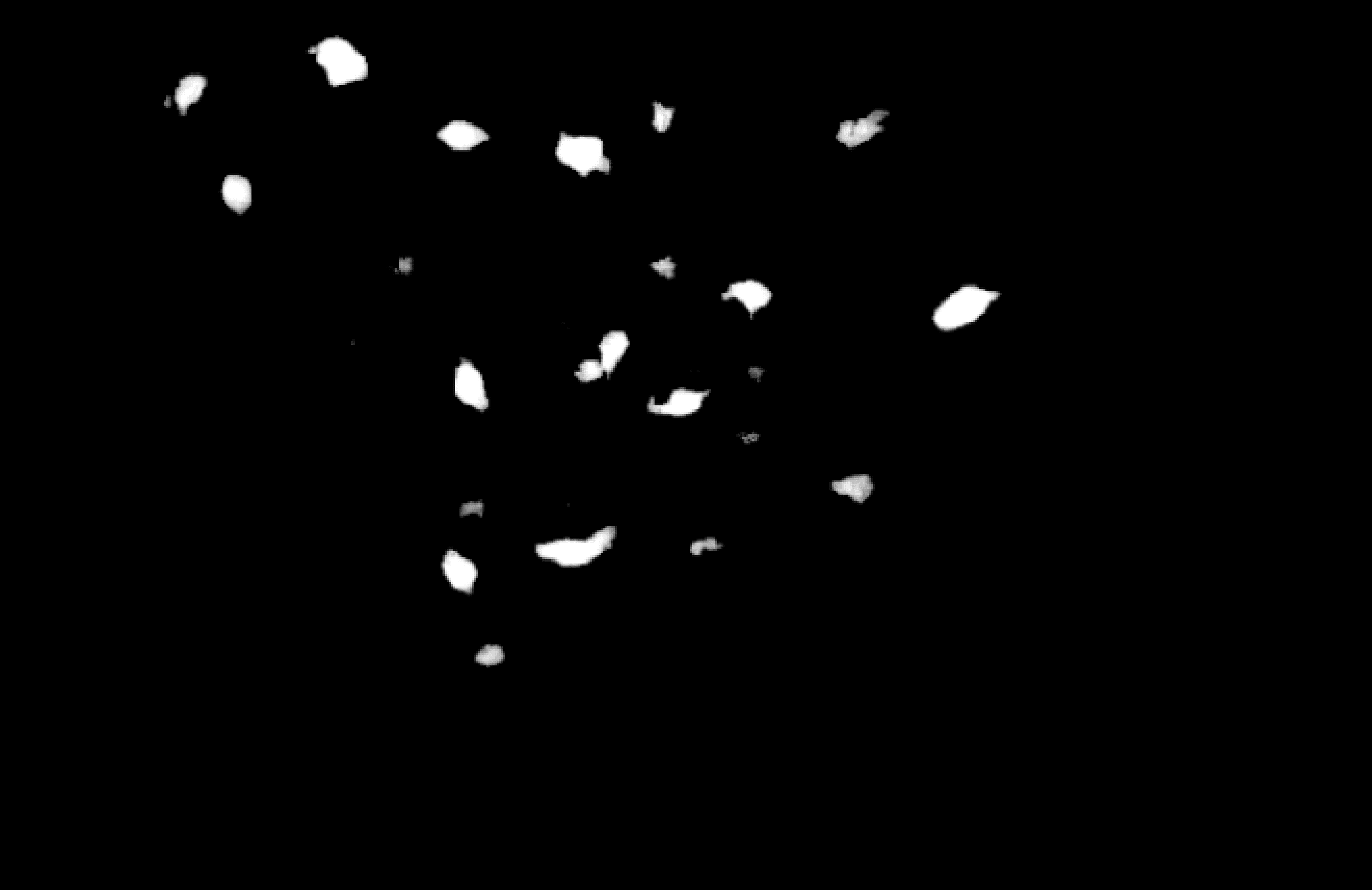}
        \caption*{NeuroSegNet}
    \end{subfigure}
    \begin{subfigure}[t]{0.16\textwidth}
        \includegraphics[width=\linewidth, height=2.8cm]{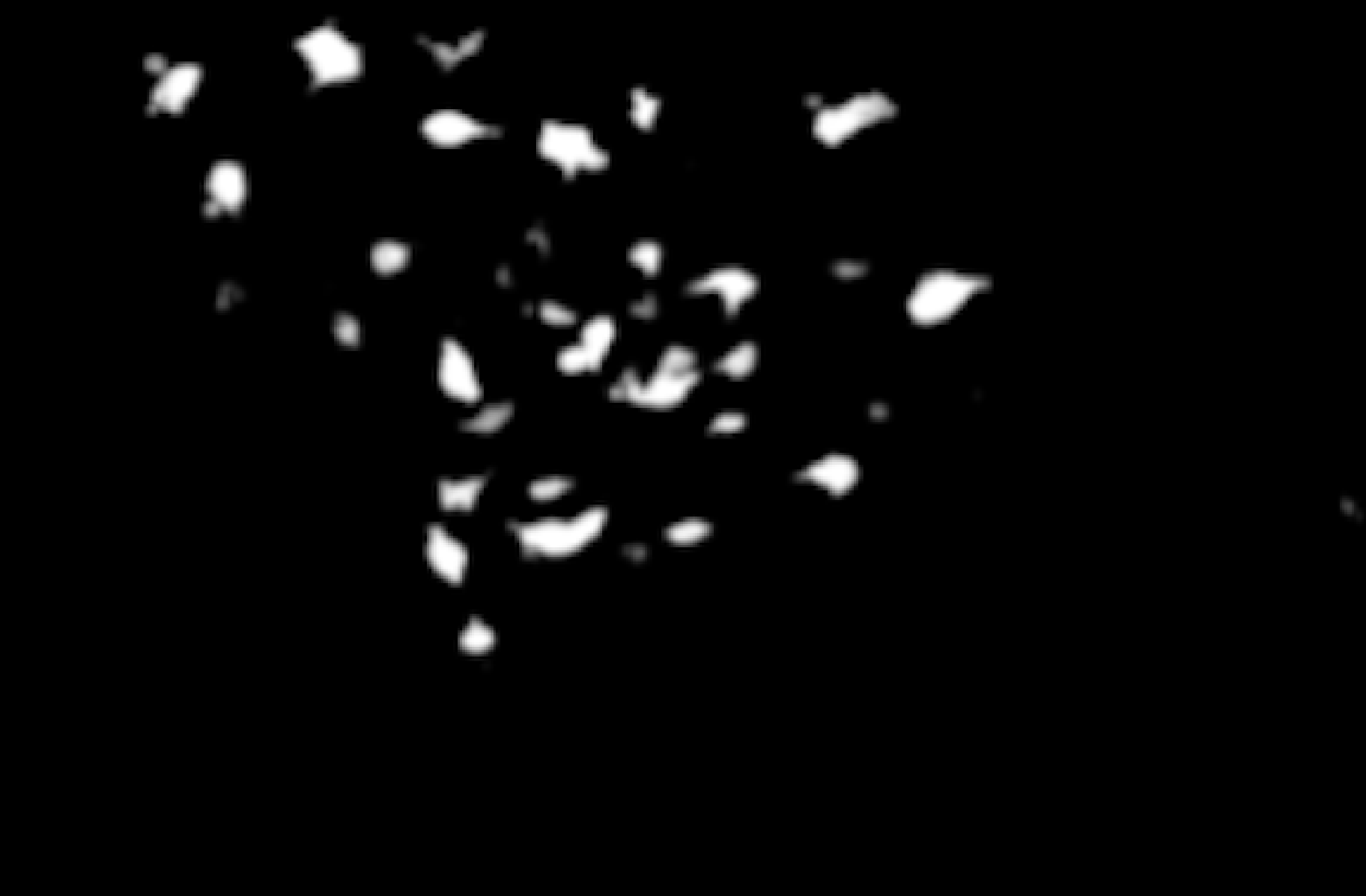}
        \caption*{IMSAHLO}
    \end{subfigure}

    % Sparse Example 1
    \vspace{2mm}
    \begin{subfigure}[t]{0.16\textwidth}
        \includegraphics[width=\linewidth, height=2.8cm]{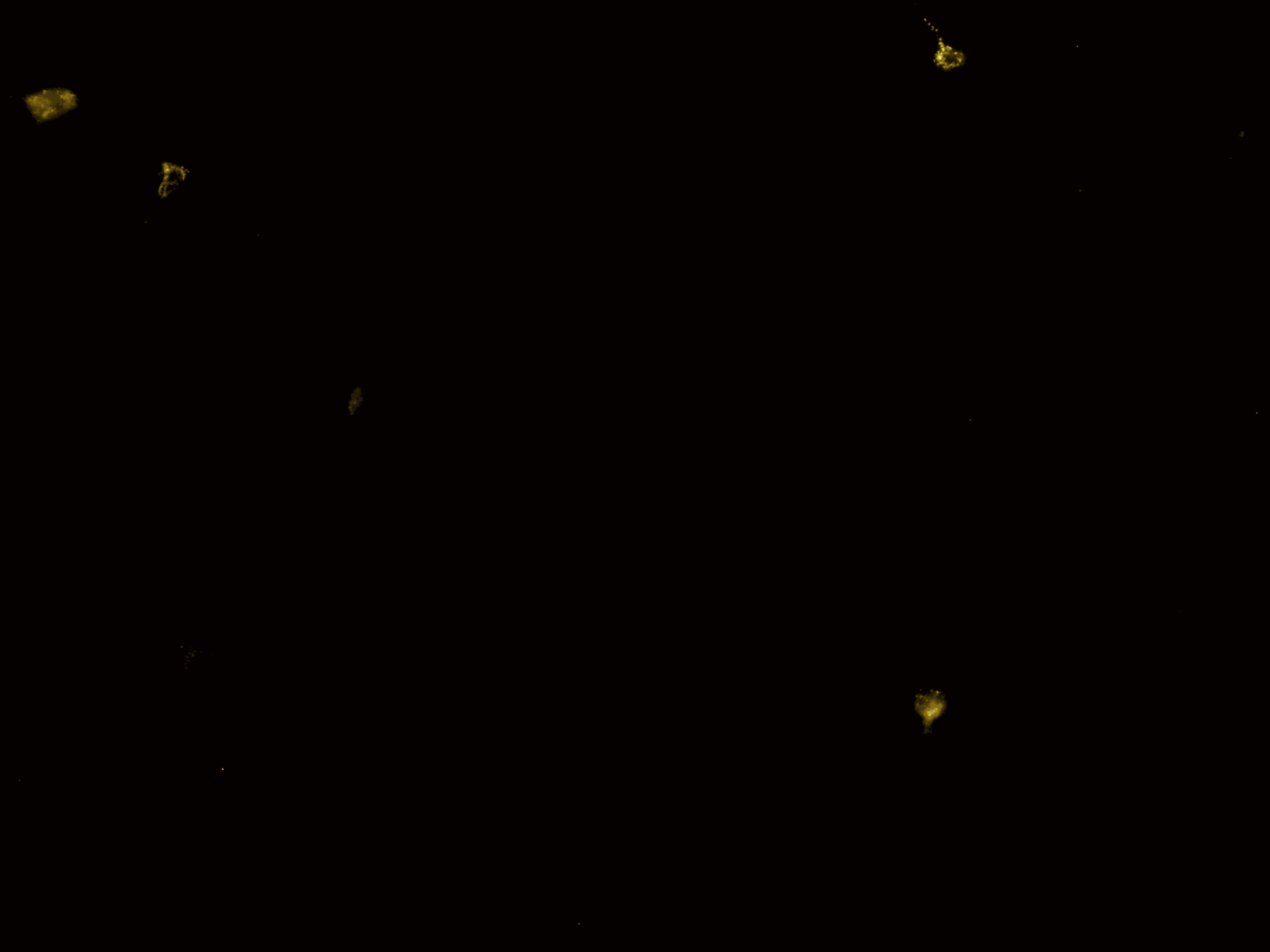}
        \caption*{Input (Sparse 1)}
    \end{subfigure}
    \begin{subfigure}[t]{0.16\textwidth}
        \includegraphics[width=\linewidth, height=2.8cm]{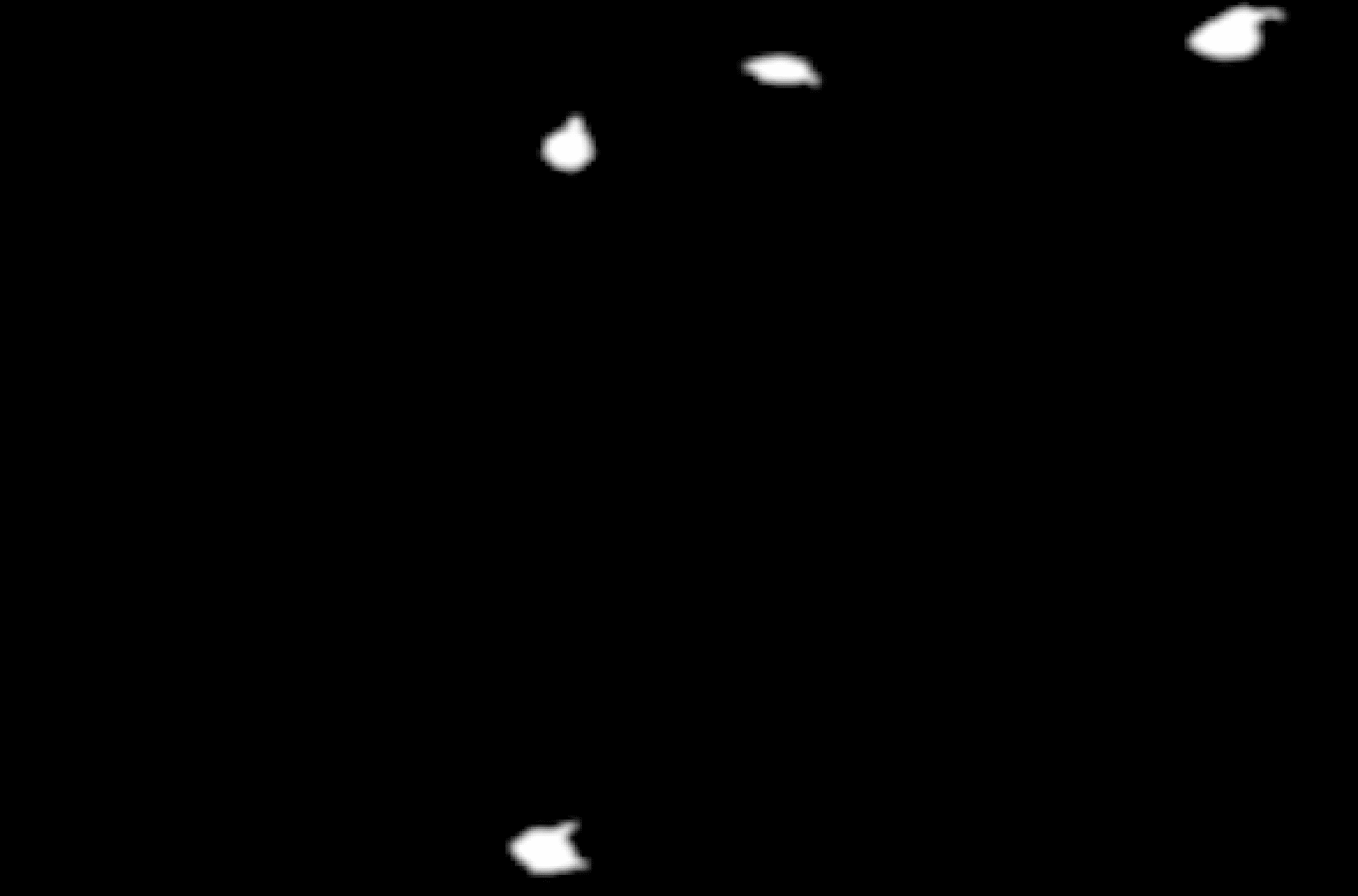}
        \caption*{GT}
    \end{subfigure}
    \begin{subfigure}[t]{0.16\textwidth}
        \includegraphics[width=\linewidth, height=2.8cm]{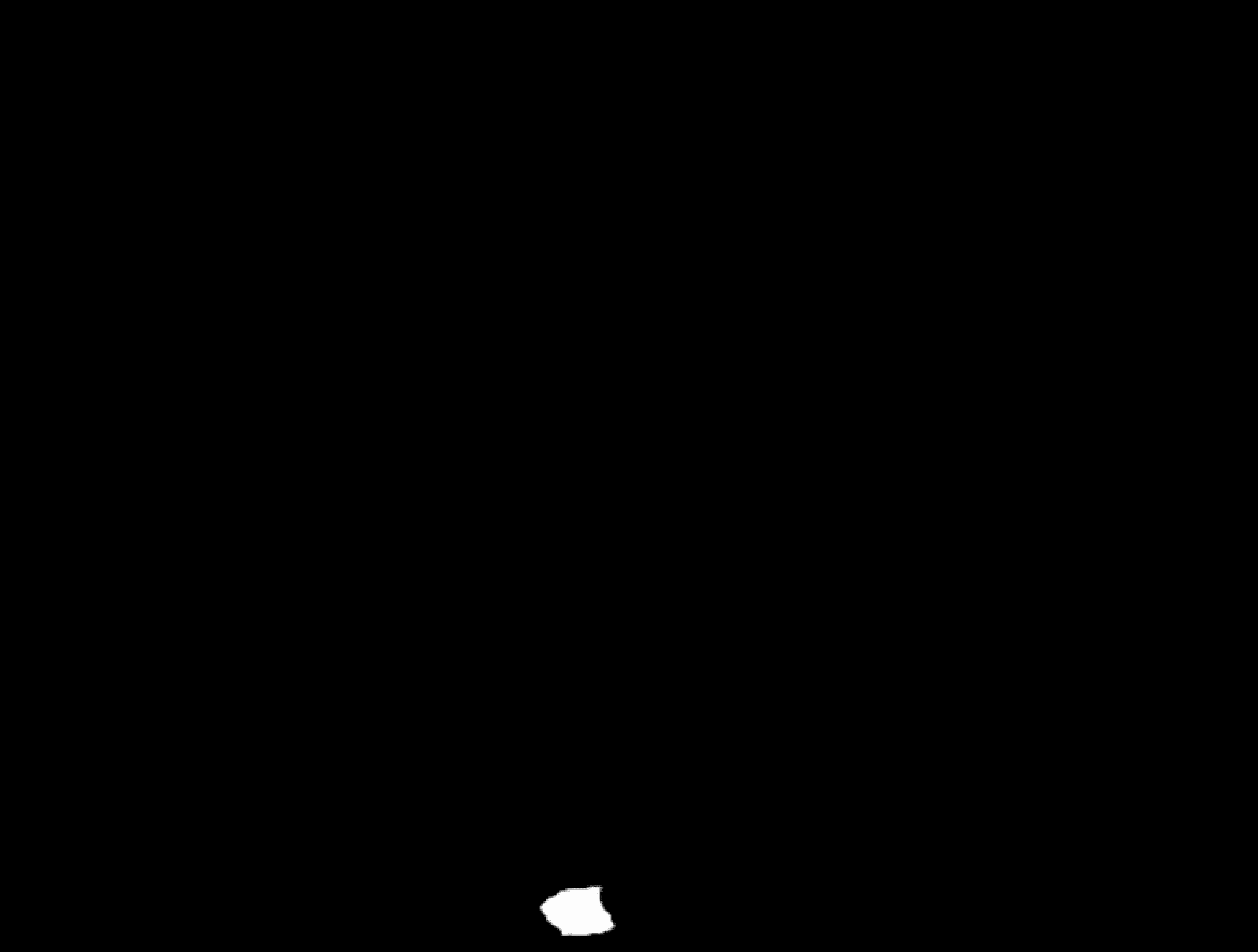}
        \caption*{TransUNet}
    \end{subfigure}
    \begin{subfigure}[t]{0.16\textwidth}
        \includegraphics[width=\linewidth, height=2.8cm]{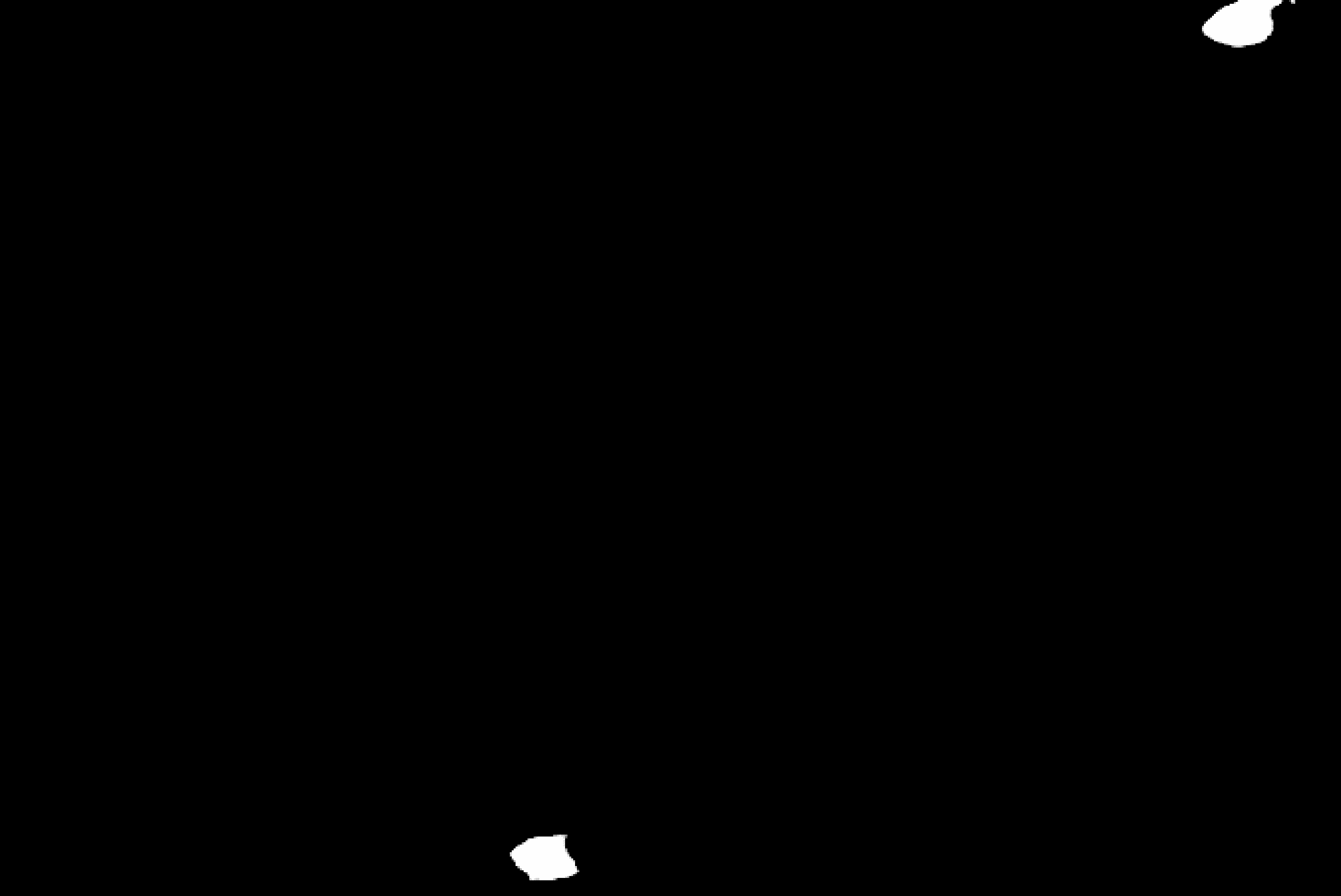}
        \caption*{DTASUNet}
    \end{subfigure}
    \begin{subfigure}[t]{0.16\textwidth}
        \includegraphics[width=\linewidth, height=2.8cm]{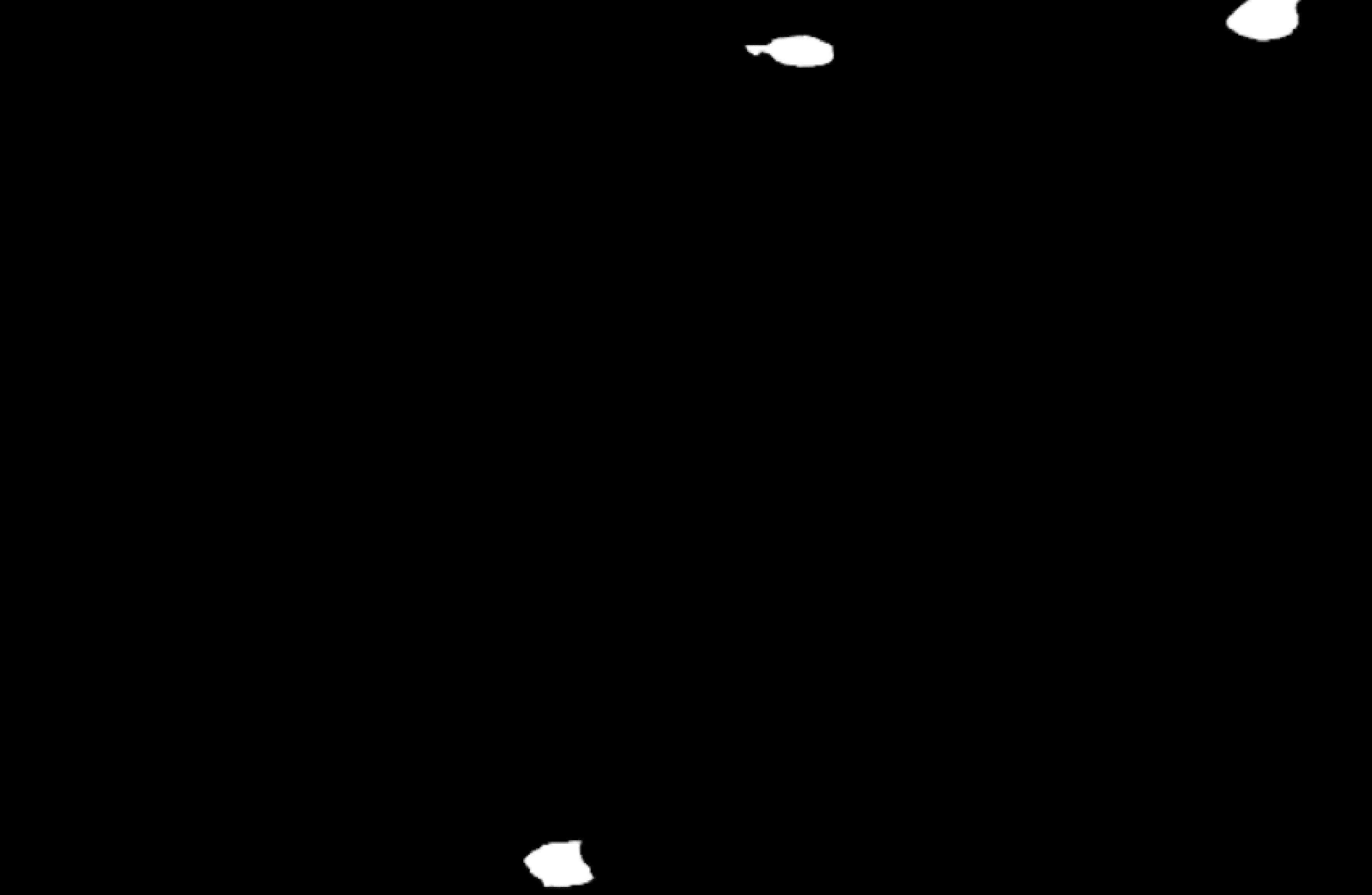}
        \caption*{NeuroSegNet}
    \end{subfigure}
    \begin{subfigure}[t]{0.16\textwidth}
        \includegraphics[width=\linewidth, height=2.8cm]{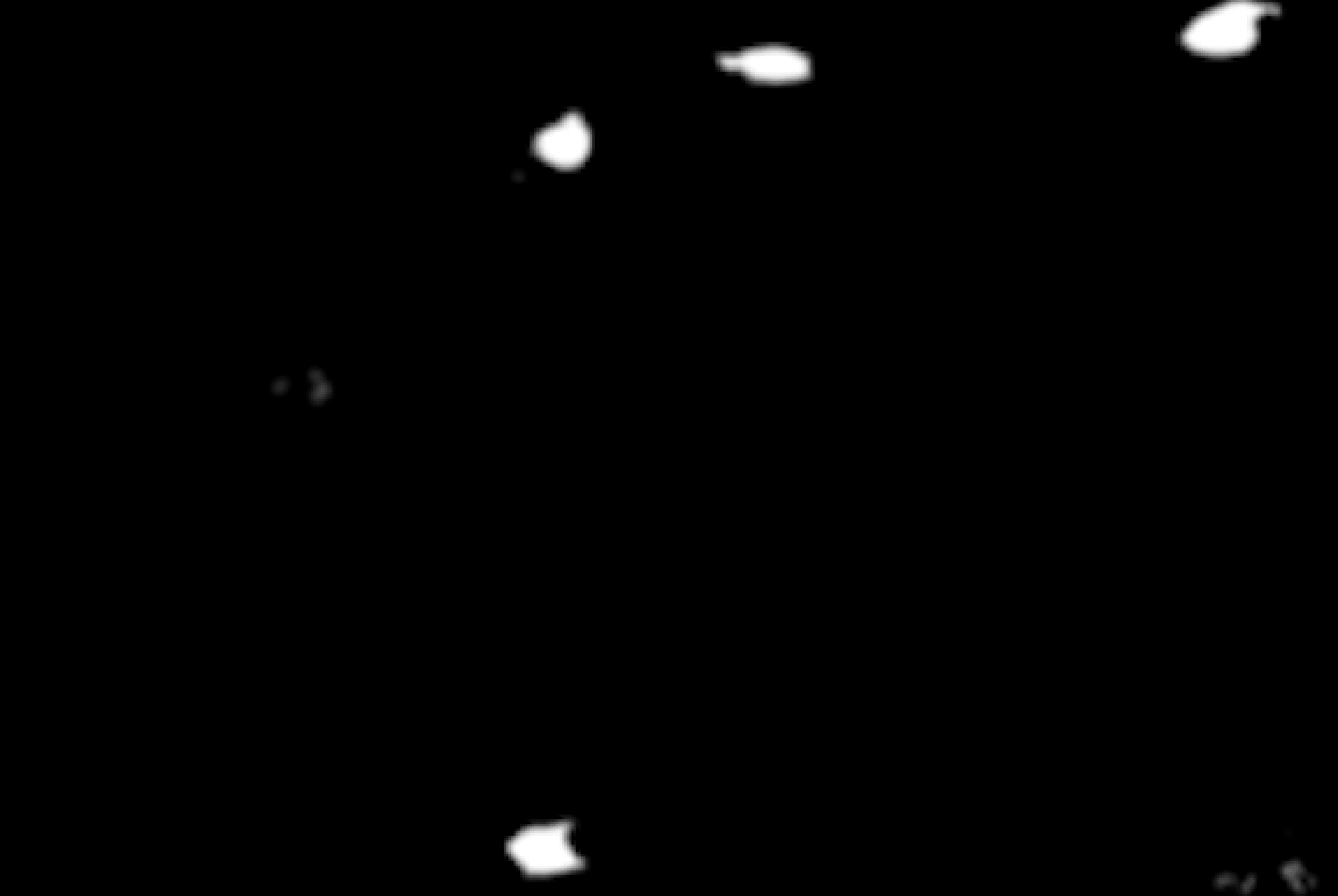}
        \caption*{IMSAHLO}
    \end{subfigure}

    % Sparse Example 2
    \vspace{2mm}
    \begin{subfigure}[t]{0.16\textwidth}
        \includegraphics[width=\linewidth, height=2.8cm]{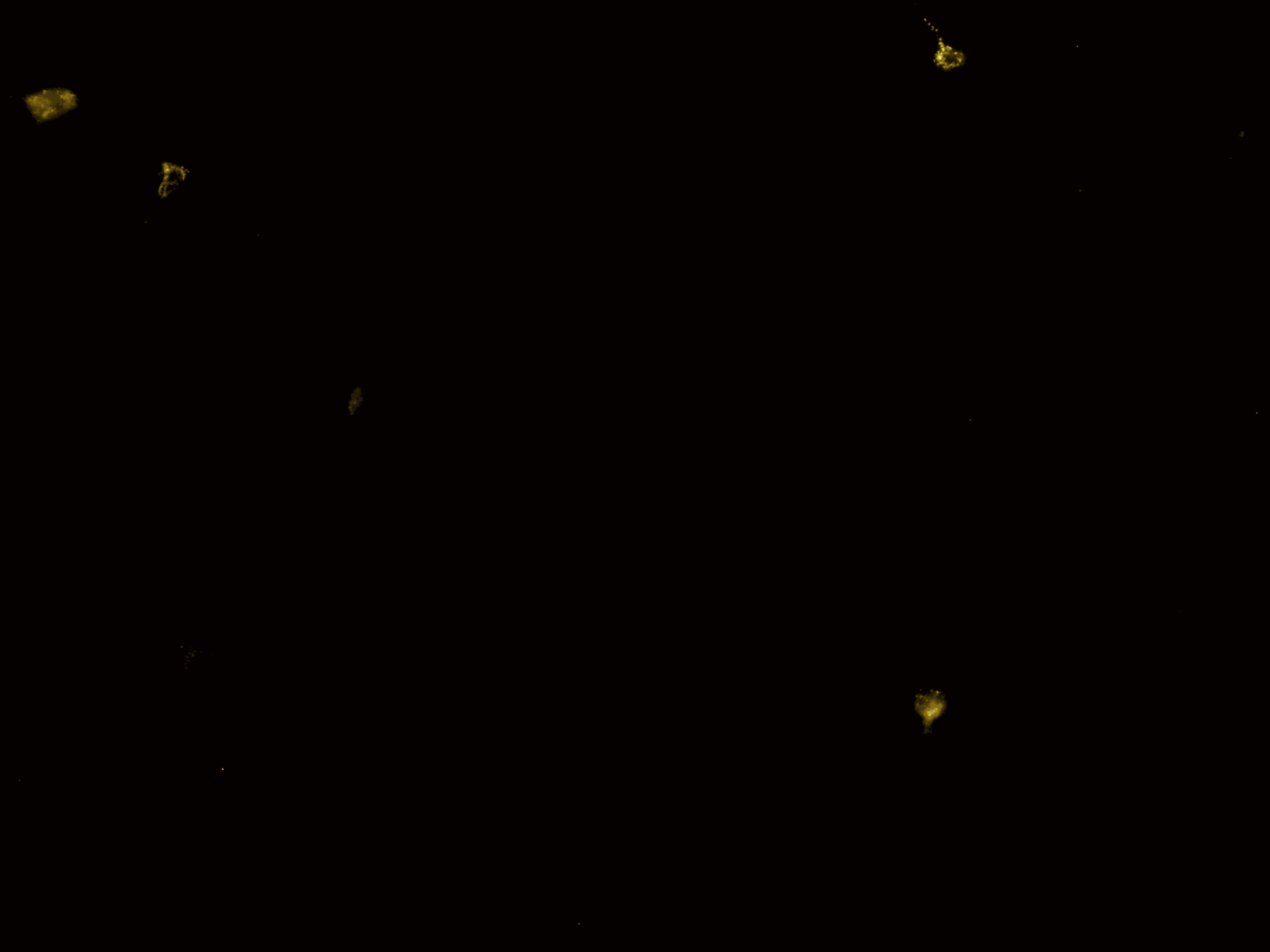}
        \caption*{Input (Sparse 2)}
    \end{subfigure}
    \begin{subfigure}[t]{0.16\textwidth}
        \includegraphics[width=\linewidth, height=2.8cm]{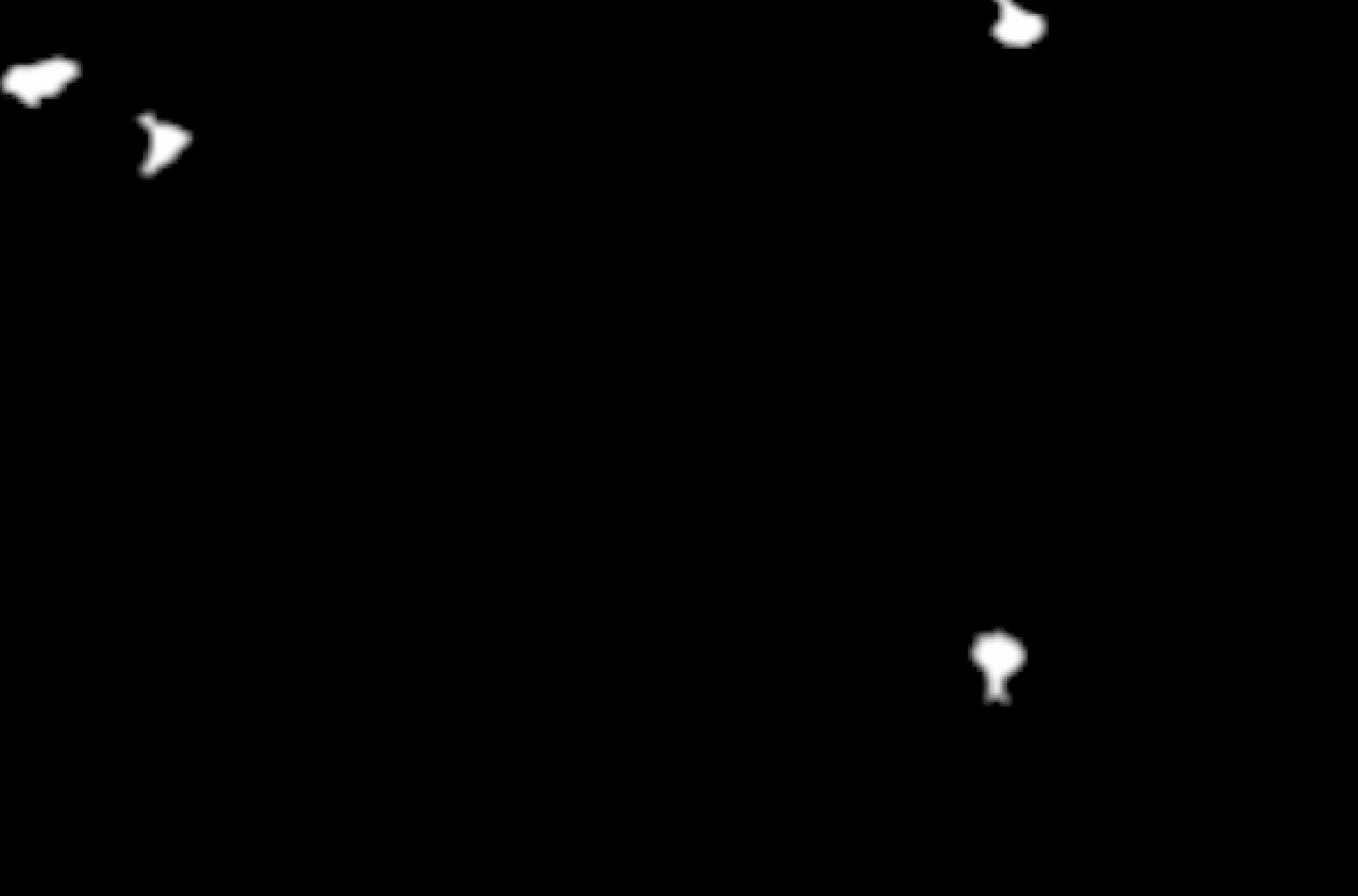}
        \caption*{GT}
    \end{subfigure}
    \begin{subfigure}[t]{0.16\textwidth}
        \includegraphics[width=\linewidth, height=2.8cm]{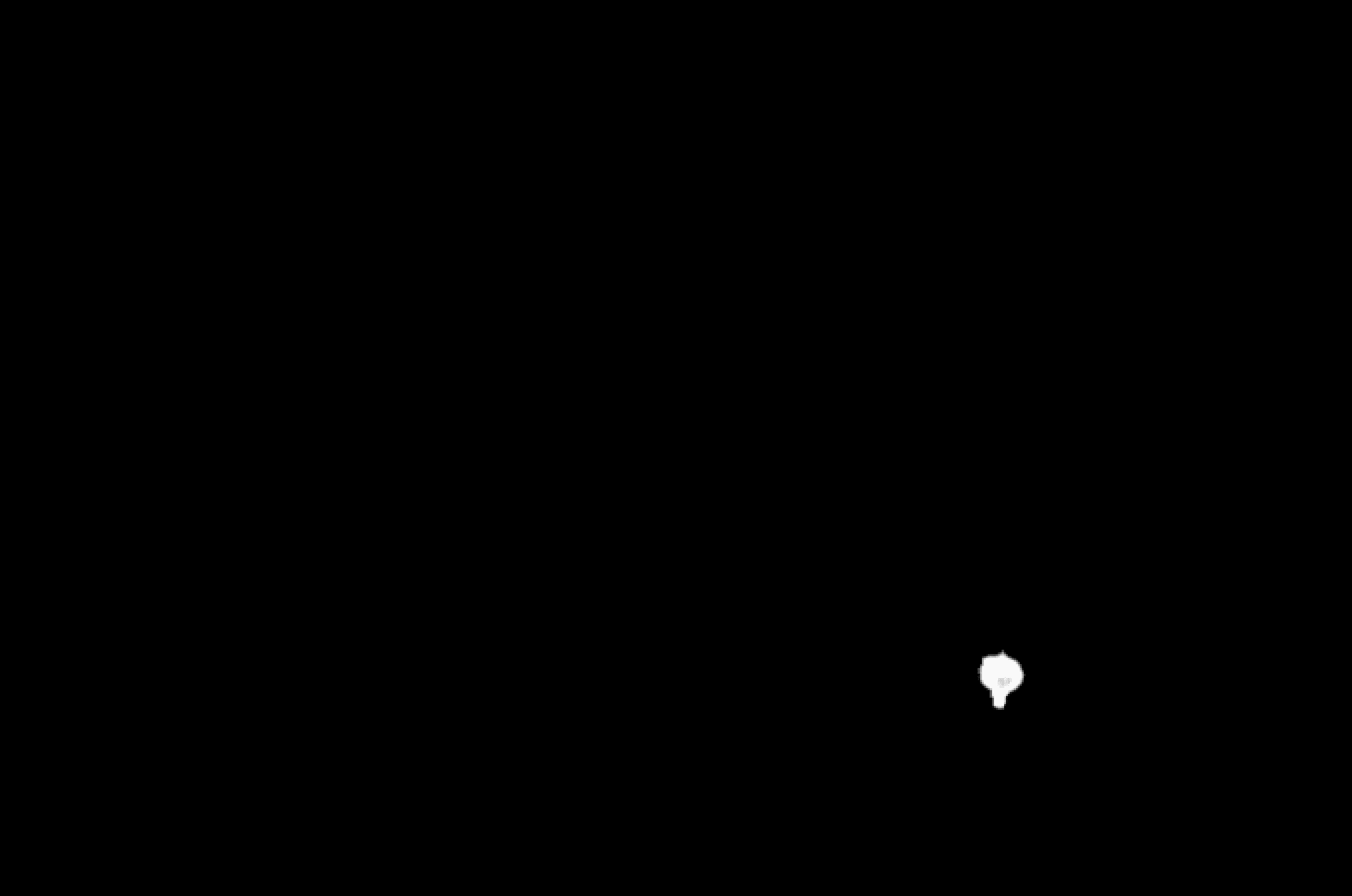}
        \caption*{TransUNet}
    \end{subfigure}
    \begin{subfigure}[t]{0.16\textwidth}
        \includegraphics[width=\linewidth, height=2.8cm]{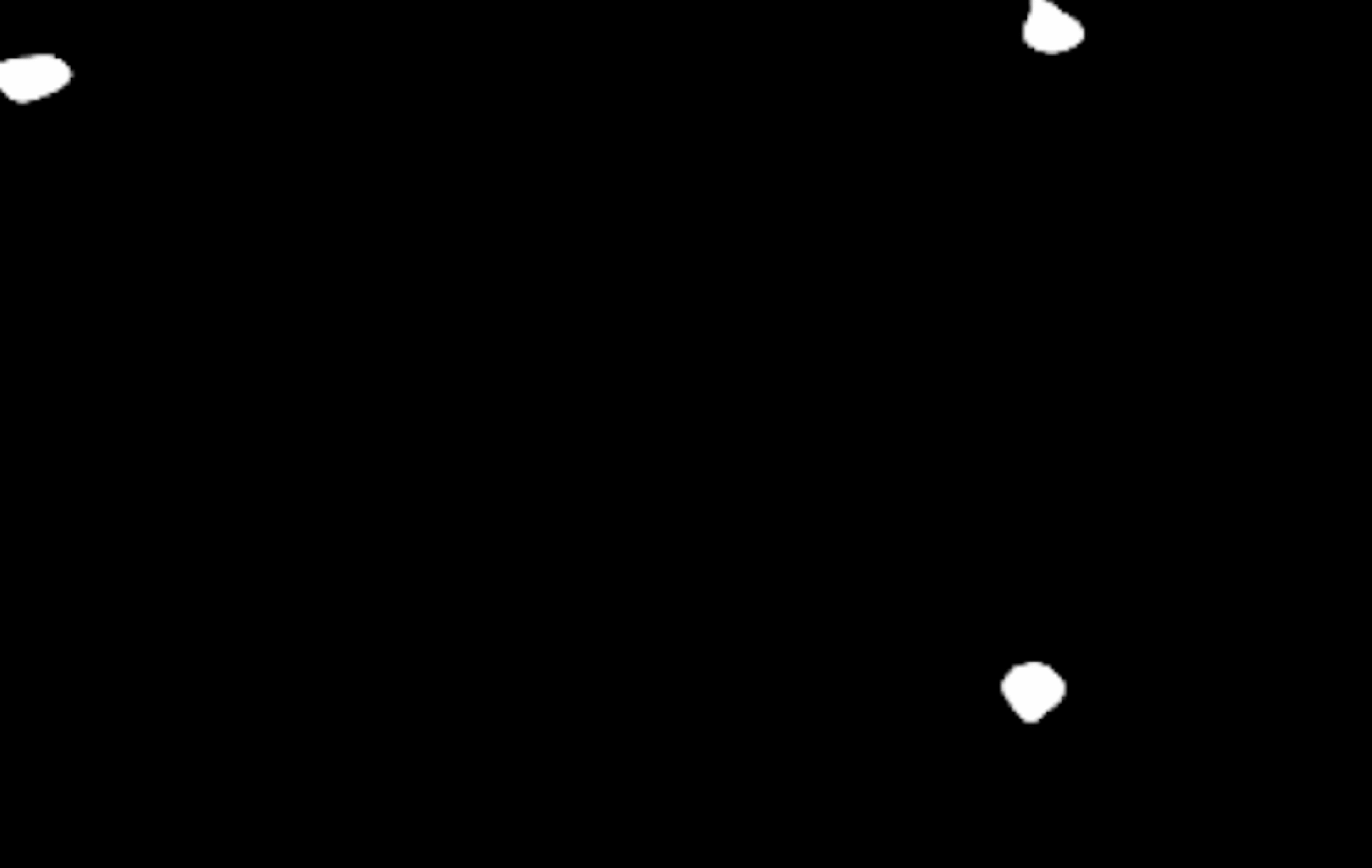}
        \caption*{DTASUNet}
    \end{subfigure}
    \begin{subfigure}[t]{0.16\textwidth}
        \includegraphics[width=\linewidth, height=2.8cm]{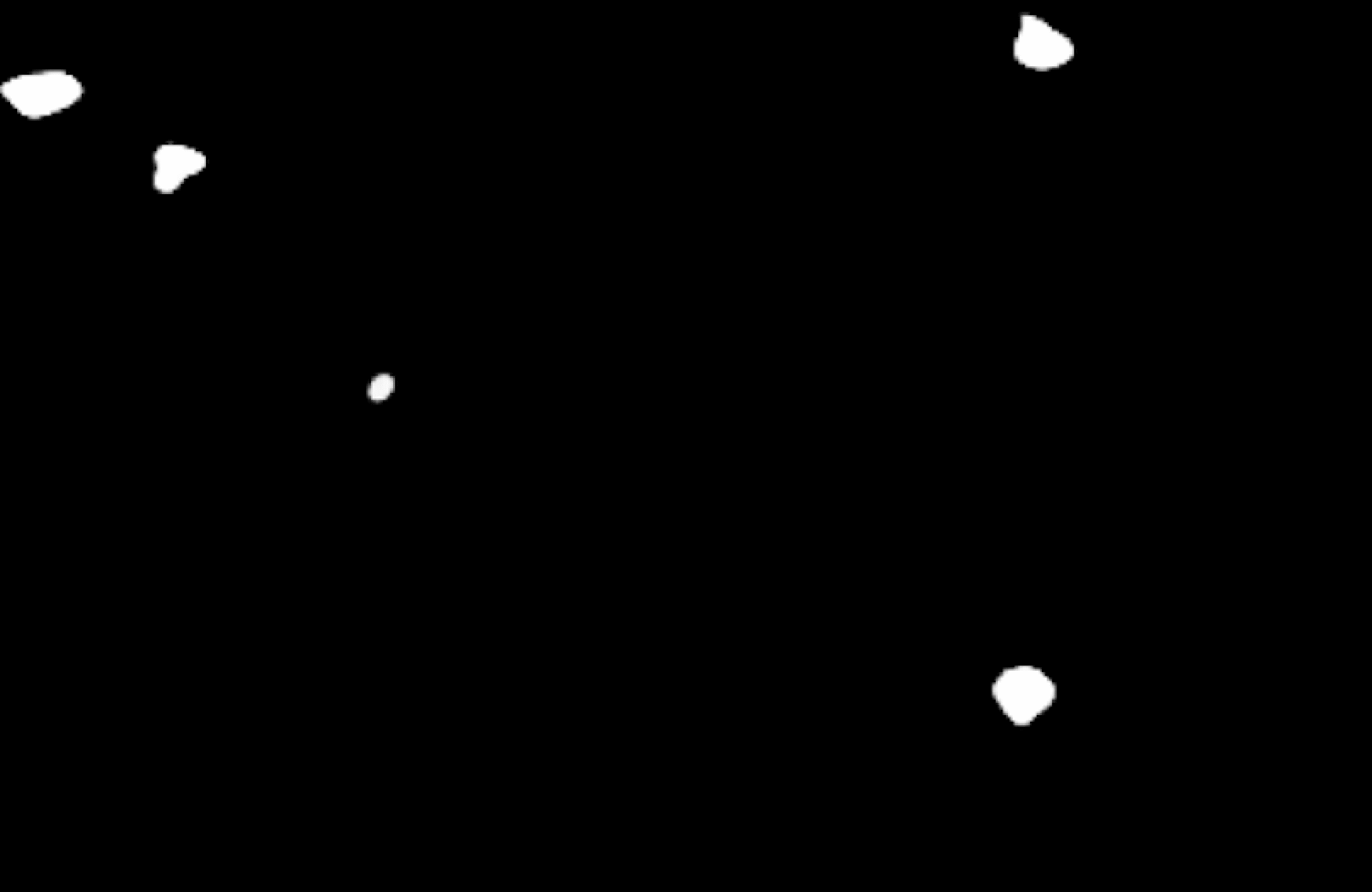}
        \caption*{NeuroSegNet}
    \end{subfigure}
    \begin{subfigure}[t]{0.16\textwidth}
        \includegraphics[width=\linewidth, height=2.8cm]{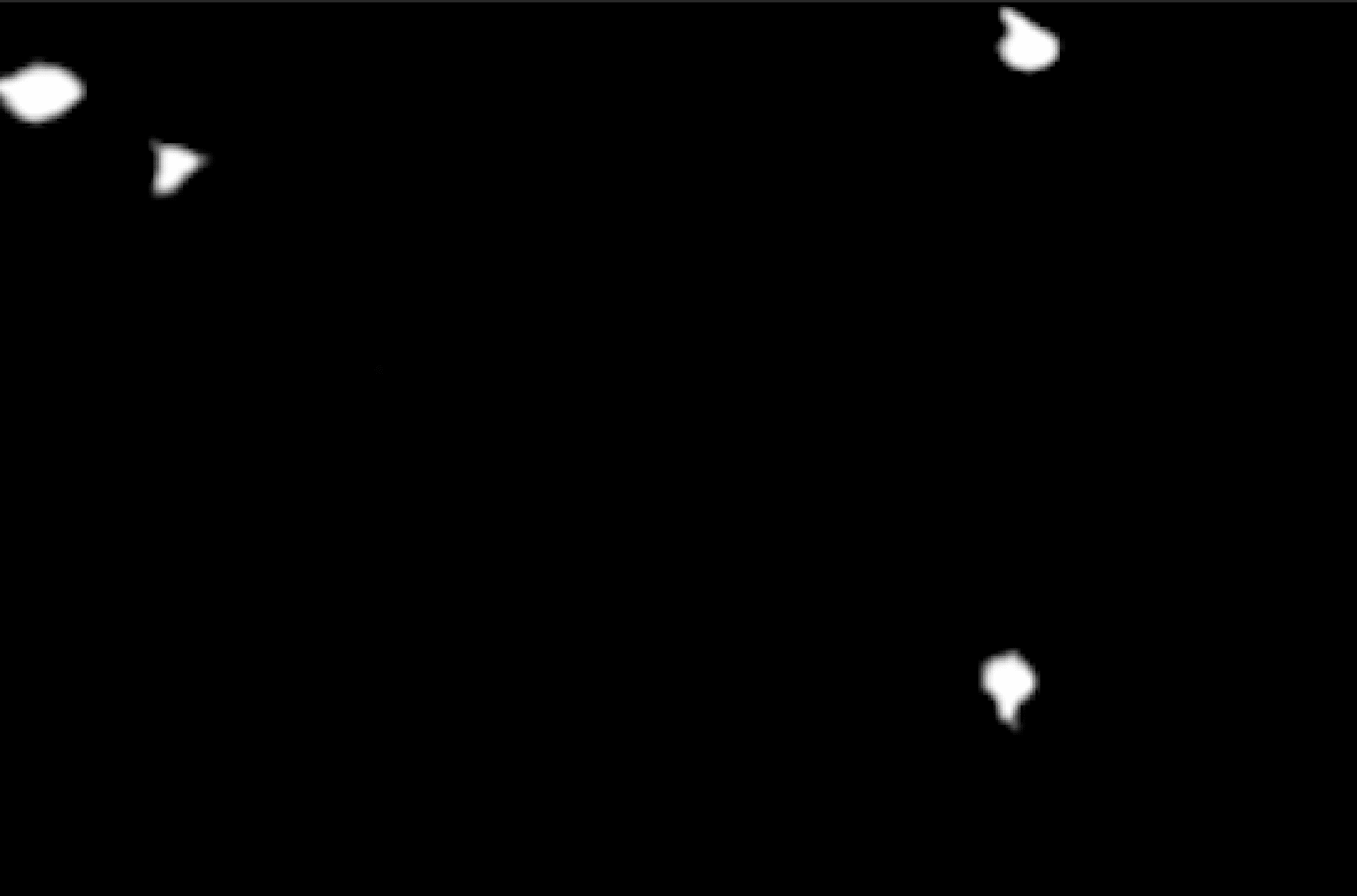}
        \caption*{IMSAHLO}
    \end{subfigure}

    \caption{Qualitative comparison of segmentation predictions across densely and sparsely packed neuronal cell regions. Each row represents a unique example showing the input image, ground truth, and predictions from TransUNet, DTASUNet, NeuroSegNet, and IMSAHLO.}
    \label{fig:qualitative_comparison}
\end{figure*}

\subsubsection{Ablation Study on Loss Function}
To analyze the contributions of the various loss terms in our proposed hybrid formulation, we conduct an ablation study on the task of neuronal cell segmentation in fluorescent microscopy images. In this work, we aim to analyze the contribution of each loss term(Tversky, Contour-Weighted Boundary, Focal, clDice) to the optimal segmentation performance when adopted in isolation or in combination as a group. Table \ref{tab:loss_ablation} summarises the results using the following evaluation metrics: Dice similarity coefficient (DSC) for measuring how much the predicted shape overlaps with the true shape, macro F1 for treating rare classes (cells) and common classes (background) as equally important, micro F1 for calculating F1 globally, and Intersection over Union (IoU) for indicating whether the segmented mask is accurate or not.
\par On the contrary, we find Tversky Loss achieves competitive recall-oriented segmentation with DSC 0.782 and IoU 0.703. This is unsurprising as it has a user-configurable bias for false negatives to help detect weak neural structures. Its boundaries, however, are not well defined. Contrastively, the Contour-Weighted Loss provides better precision and superior edge localisation but lower recall; thus it yields lower in DSC (0.738) and IoU (0.670). Focal Loss implements class imbalance by making the model more robust on hard to classify areas but it does so by the detriment of boundaries clarity and does not perform remarkably in any metric. Although clDice Loss is successful at promoting continuity through thin, tubular structures, it is only able to attain a DSC of 0.722, indicating that it is contributing to some structural regularisation rather than total region segmentation.
\par For combination of methods, Tversky + Contour increases both overlap and edge accuracy (DSC: 0.785) while Tversky + Focal trades off recall and hard pixel stressing. For three components (Tversky, Contour, and Focal), the performance achieved reaches even higher, with a DSC of 0.789 and IoU of 0.758, indicating the presence of the synergistic effect. Among these standard terms, the combination of Tversky + Contour + Focal + clDice, termed Hybrid Loss, leads to the best results in terms of DSC (0.801), Macro F1 (0.827), Micro F1 (0.833), and IoU (0.784). This result shows that a segmentation that jointly optimize region overlap, border precision, class balance and structural continuity is the most robust, in particular for the morphologically complex and class-imbalanced neuronal imaging data.
These findings confirm that each loss component tackles a different part of the segmentation challenge, and that their combination is required for high-fidelity, topology-aware biomedical segmentation. 
% \textbf{results are very good}
%\subsection{P}\label{s:catidenti}\par  \textbf{naming is odd}
\subsection{Hyperparameter Analysis}
To understand the learning behavior of IMSAHLO, we visualized the evolution of key training and validation metrics over 50 epochs, as shown in Figure \ref{fig:graphs}. The learning curves demonstrate a stable and efficient convergence, with accuracy rapidly reaching a plateau near 0.99 and loss decreasing smoothly, indicating the model generalizes effectively to unseen data without overfitting. The consistent an
d parallel improvement in both precision and recall confirms that IMSAHLO achieves a robust balance between minimizing false positives and accurately identifying true neuronal structures, which is critical for reliable segmentation.

\subsection{Qualitative Segmentation Visualization}
Figure \ref{fig:qualitative_comparison} presents quantitative segmentation on fluorescent microscopy images of brain cells on the predicted binary segmentation mask produced by the proposed method in areas with different neuronal density. We provide for each instance its input image and ground truth label as well as segmentation masks predicted by four models.
\par The first two rows show examples from its densely packed neuronal regions, where complex morphologies, cell overlaps, and varying fluorescence intensities cause significant challenging of segmentation. Nevertheless, the generated masks from IMSAHLO demonstrate faithful boundaries and topological continuity compared with the ground truth references, which outperform other models on appearance of fine contours and reduction of false-positives.
\par On the other hand, the third and fourth rows present sparsely populated neuronal regions where cells seem to be more isolated. Such cases would frequently result in false negatives or broken segmentations in the baseline models. However, IMSAHLO retains high sensitivity and structural truthfulness, accurately reproducing even the finest cell filaments and low-intensity signals, with a low level of background spilling.
\par Overall, these visual results indicate that IMSAHLO is robust and adaptable to achieve high-quality segmentation masks in both the high-density and low-density settings. The satisfactory and consistent results on these, for the most part, demanding microscopy conditions confirm beneficial generalization capability of the model and indicative for successful application to the real world biomedical problems.

\section{Conclusions and Future works} \label{s:con}
\par  
In this study, we tackled the challenging and multi-faceted problem of neuronal cell segmentation in fluorescence microscopy images that include the challenging combination of densely packed (clustered and overlapping structures), and sparse and low-contrast structures. We proposed IMSAHLO, a novel deep learning framework that takes a holistic step to co-design loss funtion and network structure. By combining Multi-Scale Dense Blocks (MSDB) for morphological diversity, Hierarchical Attention (HA) for focusing on salient features, and a carefully weighted hybrid loss function for multi-objective supervision, IMSAHLO achieves its goal of accurate cell segmentation performance.
Extensive experiments on the popular FNC dataset illustrate that IMSAHLO outperforms the existing models in regional overlap, boundary precision and class-balanced accuracy. The ablation studies empirically showed the contributions of individual architectural and loss components, and thus demonstrated that the power of the model comes from the synergy between the parts. The qualitative results also demonstrated the model could generate topologically correct and morphologically plausible segmentations whether in dense or sparse case.
This work paves the way to make segmentation models that are more adaptable and generalizable to a wide range of challenging biomedical imaging modalities. Several exciting possibilities exist for extending the IMSAHLO framework in future work. First, whereas our method is effective at disentangling most overlapping cells, explicit modeling of the instance segmentation (such as predicting an instance-specific contour or embedding) may further boost performance in the most crowded regions. Second, the framework can be generalized and assessed for 3D volumetric segmentation from confocal or light-sheet microscopy data, where the retention of 3D topology is of even greater importance. Finally, investigating semi-supervised or self-supervised learning approaches would help prefix the necessity for large and weekly annotated datasets and as a result make such powerful models as IMSAHLO more widely applicable to a broader palette of neurobiological research applications.

\bibliographystyle{IEEEtran}
\bibliography{tBM_Final}

@inproceedings{ronneberger2015u,
  title={U-net: Convolutional networks for biomedical image segmentation},
  author={Ronneberger, Olaf and Fischer, Philipp and Brox, Thomas},
  booktitle={Medical image computing and computer-assisted intervention--MICCAI 2015: 18th international conference, Munich, Germany, October 5-9, 2015, proceedings, part III 18},
  pages={234--241},
  year={2015},
  organization={Springer International Publishing}
}

@inproceedings{zhou2018unet++,
  title={Unet++: A nested U-Net architecture for medical image segmentation},
  author={Zhou, Zongwei and Siddiquee, Md Mahfuzur Rahman and Tajbakhsh, Nima and Liang, Jianming},
  booktitle={Deep learning in medical image analysis and multimodal learning for clinical decision support: 4th international workshop, DLMIA 2018, and 8th international workshop, ML-CDS 2018, held in conjunction with MICCAI 2018, Granada, Spain, September 20, 2018, proceedings 4},
  pages={3--11},
  year={2018},
  organization={Springer International Publishing}
}

@article{soydaner2022attention,
  title={Attention mechanism in neural networks: where it comes and where it goes},
  author={Soydaner, Derya},
  journal={Neural Computing and Applications},
  volume={34},
  number={16},
  pages={13371--13385},
  year={2022},
  publisher={Springer}
}

@inproceedings{vaswani2017attention,
  title={Attention is all you need},
  author={Vaswani, Ashish and Shazeer, Noam and Parmar, Niki and Uszkoreit, Jakob and Jones, Llion and Gomez, Aidan N and Kaiser, {\L}ukasz and Polosukhin, Illia},
  booktitle={Advances in neural information processing systems 30},
  pages={5998--6008},
  year={2017}
}

@inproceedings{peng2025u,
  title={U-net v2: Rethinking the skip connections of u-net for medical image segmentation},
  author={Peng, Yaopeng and Chen, Danny Z and Sonka, Milan},
  booktitle={2025 IEEE 22nd International Symposium on Biomedical Imaging (ISBI)},
  pages={1--5},
  year={2025},
  organization={IEEE}
}

@inproceedings{hu2018squeeze,
  title={Squeeze-and-excitation networks},
  author={Hu, Jie and Shen, Li and Sun, Gang},
  booktitle={Proceedings of the IEEE conference on computer vision and pattern recognition},
  pages={7132--7141},
  year={2018}
}

@article{huang2017multi,
  title={Multi-scale dense convolutional networks for efficient prediction},
  author={Huang, Gao and Chen, Danlu and Li, Tianhong and Wu, Felix and Van Der Maaten, Laurens and Weinberger, Kilian Q},
  journal={arXiv preprint arXiv:1703.09844},
  volume={2},
  number={2},
  year={2017}
}

@inproceedings{chang2019multi,
  title={Multi-scale dense network for single-image super-resolution},
  author={Chang, Chia-Yang and Chien, Shao-Yi},
  booktitle={ICASSP 2019-2019 IEEE International Conference on Acoustics, Speech and Signal Processing (ICASSP)},
  pages={1742--1746},
  year={2019},
  organization={IEEE}
}

@article{Ding2019HierarchicalAN,
  title={Hierarchical attention networks for medical image segmentation},
  author={Fei Ding and Gang Yang and Jinlu Liu and Jun Wu and Dayong Ding and Jie Xv and Gangwei Cheng and Xirong Li},
  journal={arXiv preprint arXiv:1911.08777},
  year={2019}
}

@inproceedings{Wang2020ECANetEC,
  title={ECA-Net: Efficient channel attention for deep convolutional neural networks},
  author={Qilong Wang and Banggu Wu and Pengfei Zhu and Peihua Li and Wangmeng Zuo and Qinghua Hu},
  booktitle={Proceedings of the IEEE/CVF conference on computer vision and pattern recognition},
  pages={11534--11542},
  year={2020}
}

@article{Xie2023ResidualTW,
  title={Residual: Transformer with dual residual connections},
  author={Shufang Xie and Huishuai Zhang and Junliang Guo and Xu Tan and Jiang Bian and Hany Hassan Awadalla and Arul Menezes and Tao Qin and Rui Yan},
  journal={arXiv preprint arXiv:2304.14802},
  year={2023}
}

@article{Wu2021UNetCW,
  title={U-Net combined with multi-scale attention mechanism for liver segmentation in CT images},
  author={Jiawei Wu and Shengqiang Zhou and Songlin Zuo and Yiyin Chen and Weiqin Sun and Jiang Luo and Jiantuan Duan and Hui Wang and Deguang Wang},
  journal={BMC Medical Informatics and Decision Making},
  volume={21},
  pages={1--12},
  year={2021}
}

@article{Krikid2024StateofthetotDL,
  title={State-of-the-Art Deep Learning Methods for Microscopic Image Segmentation: Applications to Cells, Nuclei, and Tissues},
  author={Fatma Krikid and others},
  journal={Journal of Imaging},
  volume={10},
  number={12},
  pages={311},
  year={2024}
}

@article{Wang2024MultimodalPA,
  title={Multimodal parallel attention network for medical image segmentation},
  author={Zhibing Wang and Wenmin Wang and Nannan Li and Shenyong Zhang and Qi Chen and Zhe Jiang},
  journal={Image and Vision Computing},
  volume={147},
  pages={105069},
  year={2024}
}

@inproceedings{Zhang2022SemanticSM,
  title={Semantic Segmentation Model of Fluorescent Neuronal Cells in Mouse Brain Slices Under Few Samples},
  author={Zehua Zhang and Bailing Liu and Gaohao Zhou},
  booktitle={Proceedings of the 2022 6th International Conference on Virtual and Augmented Reality Simulations},
  pages={64--70},
  year={2022}
}

@inproceedings{Hiramatsu2018CellIS,
  title={Cell image segmentation by integrating multiple CNNs},
  author={Yuki Hiramatsu and Kazuhiro Hotta and Ayako Imanishi and Michiyuki Matsuda and Kenta Terai},
  booktitle={Proceedings of the IEEE Conference on Computer Vision and Pattern Recognition Workshops},
  pages={2205--2211},
  year={2018}
}

@article{Xu2024CrossDomainAG,
  title={Cross-Domain Attention-Guided Generative Data Augmentation for Medical Image Analysis with Limited Data},
  author={Zhenghua Xu and others},
  journal={Computers in Biology and Medicine},
  volume={168},
  pages={107744},
  year={2024}
}

@article{Huang2024ChannelPC,
  title={Channel prior convolutional attention for medical image segmentation},
  author={Hejun Huang and Zuguo Chen and Ying Zou and Ming Lu and Chaoyang Chen and Youzhi Song and Hongqiang Zhang and Feng Yan},
  journal={Computers in Biology and Medicine},
  volume={178},
  pages={108784},
  year={2024}
}

@article{Oktay2018AttentionUL,
  title={Attention u-net: Learning where to look for the pancreas},
  author={Ozan Oktay and Jo Schlemper and Loic Le Folgoc and Matthew Lee and Mattias Heinrich and Kazunari Misawa and Kensaku Mori and others},
  journal={arXiv preprint arXiv:1804.03999},
  year={2018}
}

@inproceedings{Chen2018EncoderdecoderWA,
  title={Encoder-decoder with atrous separable convolution for semantic image segmentation},
  author={Liang-Chieh Chen and Yukun Zhu and George Papandreou and Florian Schroff and Hartwig Adam},
  booktitle={Proceedings of the European conference on computer vision (ECCV)},
  pages={801--818},
  year={2018}
}

@article{Mehta2021MobilevitLG,
  title={Mobilevit: light-weight, general-purpose, and mobile-friendly vision transformer},
  author={Sachin Mehta and Mohammad Rastegari},
  journal={arXiv preprint arXiv:2110.02178},
  year={2021}
}

@article{Chen2021TransunetTM,
  title={Transunet: Transformers make strong encoders for medical image segmentation},
  author={Jieneng Chen and Yongyi Lu and Qihang Yu and Xiangde Luo and Ehsan Adeli and Yan Wang and Le Lu and Alan L. Yuille and Yuyin Zhou},
  journal={arXiv preprint arXiv:2102.04306},
  year={2021}
}

@inproceedings{Valanarasu2022UNEXTMR,
  title={Unext: Mlp-based rapid medical image segmentation network},
  author={Jeya Maria Jose Valanarasu and Vishal M. Patel},
  booktitle={International conference on medical image computing and computer-assisted intervention},
  pages={23--33},
  year={2022},
  publisher={Cham: Springer Nature Switzerland}
}

@article{Ma2024DTASUNetAL,
  title={DTASUnet: a local and global dual transformer with the attention supervision U-network for brain tumor segmentation},
  author={Bo Ma and Qian Sun and Ze Ma and Baosheng Li and Qiang Cao and Yungang Wang and Gang Yu},
  journal={Scientific Reports},
  volume={14},
  number={1},
  pages={28379},
  year={2024}
}

@inproceedings{Huang2024ContourweightedLF,
  title={Contour-weighted loss for class-imbalanced image segmentation},
  author={Zhengyong Huang and Yao Sui},
  booktitle={2024 IEEE International Conference on Image Processing (ICIP)},
  pages={3084--3090},
  year={2024},
  organization={IEEE}
}

@inproceedings{Abraham2019ANF,
  title={A novel focal tversky loss function with improved attention u-net for lesion segmentation},
  author={Nabila Abraham and Naimul Mefraz Khan},
  booktitle={2019 IEEE 16th international symposium on biomedical imaging (ISBI 2019)},
  pages={683--687},
  year={2019},
  organization={IEEE}
}

@inproceedings{Salehi2017TverskyLF,
  title={Tversky loss function for image segmentation using 3D fully convolutional deep networks},
  author={Seyed Sadegh Mohseni Salehi and Deniz Erdogmus and Ali Gholipour},
  booktitle={International workshop on machine learning in medical imaging},
  pages={379--387},
  year={2017},
  publisher={Cham: Springer International Publishing}
}

@inproceedings{Jadon2020ASO,
  title={A survey of loss functions for semantic segmentation},
  author={Shruti Jadon},
  booktitle={2020 IEEE conference on computational intelligence in bioinformatics and computational biology (CIBCB)},
  pages={1--7},
  year={2020},
  organization={IEEE}
}

@article{Yeung2022UnifiedFL,
  title={Unified focal loss: Generalising dice and cross entropy-based losses to handle class imbalanced medical image segmentation},
  author={Michael Yeung and Evis Sala and Carola-Bibiane Schönlieb and Leonardo Rundo},
  journal={Computerized Medical Imaging and Graphics},
  volume={95},
  pages={102026},
  year={2022}
}

@inproceedings{Sudre2017GeneralisedDO,
  title={Generalised dice overlap as a deep learning loss function for highly unbalanced segmentations},
  author={Carole H. Sudre and Wenqi Li and Tom Vercauteren and Sebastien Ourselin and M. Jorge Cardoso},
  booktitle={Deep Learning in Medical Image Analysis and Multimodal Learning for Clinical Decision Support: Third International Workshop, DLMIA 2017, and 7th International Workshop, ML-CDS 2017, Held in Conjunction with MICCAI 2017, Québec City, QC, Canada, September 14, Proceedings 3},
  pages={240--248},
  year={2017},
  publisher={Springer International Publishing}
}

@inproceedings{Shi2024CenterlineBD,
  title={Centerline boundary dice loss for vascular segmentation},
  author={Pengcheng Shi and Jiesi Hu and Yanwu Yang and Zilve Gao and Wei Liu and Ting Ma},
  booktitle={International Conference on Medical Image Computing and Computer-Assisted Intervention},
  pages={46--56},
  year={2024},
  publisher={Cham: Springer Nature Switzerland}
}

@article{Zhu2022A_Compound_Loss_Function,
  title={A compound loss function with shape aware weight map for microscopy cell segmentation},
  author={Yanming Zhu and Xuefei Yin and Erik Meijering},
  journal={IEEE Transactions on Medical Imaging},
  volume={42},
  number={5},
  pages={1278--1288},
  year={2022}
}

@article{Chen2020ContourawareLB,
  title={Contour-aware loss: Boundary-aware learning for salient object segmentation},
  author={Zixuan Chen and Huajun Zhou and Jianhuang Lai and Lingxiao Yang and Xiaohua Xie},
  journal={IEEE Transactions on Image Processing},
  volume={30},
  pages={431--443},
  year={2020}
}

@article{Zhou2025BoundaryawareAC,
  title={Boundary-aware and cross-modal fusion network for enhanced multi-modal brain tumor segmentation},
  author={Tongxue Zhou},
  journal={Pattern Recognition},
  volume={165},
  pages={111637},
  year={2025}
}

@inproceedings{Clissa2023OptimizingDL,
  title={Optimizing deep learning models for cell recognition in fluorescence microscopy: The impact of loss functions on performance and generalization},
  author={Luca Clissa and Antonio Macaluso and Antonio Zoccoli},
  booktitle={International Conference on Image Analysis and Processing},
  pages={179--190},
  year={2023},
  publisher={Cham: Springer Nature Switzerland}
}

@article{Lu2025SteerablePW,
  title={Steerable Pyramid Weighted Loss: Multi-Scale Adaptive Weighting for Semantic Segmentation},
  author={Renhao Lu},
  journal={arXiv preprint arXiv:2503.06604},
  year={2025}
}

@article{Terven2023LossFA,
  title={Loss functions and metrics in deep learning},
  author={Juan Terven and Diana M. Cordova-Esparza and Alfonso Ramirez-Pedraza and Edgar A. Chavez-Urbiola and Julio A. Romero-Gonzalez},
  journal={arXiv preprint arXiv:2307.02694},
  year={2023}
}

@inproceedings{Kervadec2019BoundaryLF,
  title={Boundary loss for highly unbalanced segmentation},
  author={Hoel Kervadec and Jihene Bouchtiba and Christian Desrosiers and Eric Granger and Jose Dolz and Ismail Ben Ayed},
  booktitle={International conference on medical imaging with deep learning},
  pages={285--296},
  year={2019},
  organization={PMLR}
}

@article{Guo2022ContourLF,
  title={Contour loss for instance segmentation via k-step distance transformation image},
  author={Xiaolong Guo and Xiaosong Lan and Kunfeng Wang and Shuxiao Li},
  journal={IET Computer Vision},
  volume={16},
  number={8},
  pages={683--693},
  year={2022}
}

@article{Cheng2020HybridlossSF,
  title={Hybrid-loss supervision for deep neural network},
  author={Qishang Cheng and Hongliang Li and Qingbo Wu and King Ngi Ngan},
  journal={Neurocomputing},
  volume={388},
  pages={78--89},
  year={2020}
}

@article{Morelli2021AutomatingCC,
  title={Automating cell counting in fluorescent microscopy through deep learning with c-ResUnet},
  author={R. Morelli and Luca Clissa and R. Amici and M. Cerri and T. Hitrec and M. Luppi and L. Rinaldi and F. Squarcio and Antonio Zoccoli},
  journal={Scientific Reports},
  volume={11},
  pages={22920},
  year={2021},
  publisher={Nature Publishing Group}, 
  doi={10.1038/s41598-021-01929-5}
}

@article{haque2020deep,
  title={Deep learning approaches to biomedical image segmentation},
  author={Haque, Intisar Rizwan I. and Neubert, Jeremiah},
  journal={Informatics in Medicine Unlocked},
  volume={18},
  pages={100297},
  year={2020},
  publisher={Elsevier},
  doi={10.1016/j.imu.2020.100297}
}

@article{xu2024advances,
  title={Advances in medical image segmentation: A comprehensive review of traditional, deep learning and hybrid approaches},
  author={Xu, Yan and Quan, Rixiang and Xu, Weiting and Huang, Yi and Chen, Xiaolong and Liu, Fengyuan},
  journal={Bioengineering},
  volume={11},
  number={10},
  pages={1034},
  year={2024},
  publisher={MDPI},
  doi={10.3390/bioengineering11101034}
}

@inproceedings{roy2018recalibrating,
  author    = {Anirban Ghosh Roy and Nassir Navab and Christian Wachinger},
  title     = {Recalibrating Fully Convolutional Networks with Spatial and Channel “Squeeze and Excitation” Blocks},
  booktitle = {International Conference on Medical Image Computing and Computer-Assisted Intervention (MICCAI)},
  pages     = {202--210},
  year      = {2018},
  publisher = {Springer, Cham},
  doi       = {10.1007/978-3-030-00928-1_48},
  url       = {https://link.springer.com/chapter/10.1007/978-3-030-00928-1_48}
}

@inproceedings{lin2017focal,
  author    = {Tsung-Yi Lin and Priya Goyal and Ross Girshick and Kaiming He and Piotr Doll{\'a}r},
  title     = {Focal Loss for Dense Object Detection},
  booktitle = {Proceedings of the IEEE International Conference on Computer Vision (ICCV)},
  year      = {2017},
  pages     = {2980--2988},
  publisher = {IEEE},
  doi       = {10.1109/ICCV.2017.324},
  url       = {https://openaccess.thecvf.com/content_iccv_2017/html/Lin_Focal_Loss_for_ICCV_2017_paper.html}
}

@article{mohammed2025contrast,
  title={Contrast Limited Adaptive Local Histogram Equalization Method for Poor Contrast Image Enhancement},
  author={Mohammed, Ibrahim Majid and Isa, Nor Ashidi Mat},
  journal={IEEE Access},
  year={2025},
  publisher={IEEE}
}

@inproceedings{kimura2021understanding,
  author    = {Masanari Kimura},
  title     = {Understanding Test-Time Augmentation},
  booktitle = {International Conference on Neural Information Processing},
  pages     = {558--569},
  year      = {2021},
  publisher = {Springer International Publishing},
  address   = {Cham},
  doi       = {10.1007/978-3-030-92310-6_46},
  url       = {https://link.springer.com/chapter/10.1007/978-3-030-92310-6_46}
}

@article{nourollah2024quantifying,
  title={Quantifying morphologies of developing neuronal cells using deep learning with imperfect annotations},
  author={Nourollah, AM and Hassanpour, H and Zehtabian, A},
  journal={IBRO Neuroscience Reports},
  year={2024},
  publisher={Elsevier},
  url={https://www.sciencedirect.com/science/article/pii/S2667242123022959}
}

@article{zarb2025multimodal,
  title={Multimodal synapse analysis reveals restricted integration of transplanted neurons remodeled by TREM2},
  author={Zarb, Y and Thorwirth, M and Markkula, O and Paoli, E and others},
  journal={bioRxiv},
  year={2025},
  publisher={Cold Spring Harbor Laboratory},
  url={https://www.biorxiv.org/content/biorxiv/early/2025/01/31/2025.01.31.635250.full.pdf}
}

@article{debeuckeleer2025unbiased,
  title={Unbiased identification of cell identity in dense mixed neural cultures},
  author={De Beuckeleer, S and Van De Looverbosch, T and others},
  journal={eLife},
  year={2025},
  publisher={eLife Sciences Publications},
  url={https://elifesciences.org/articles/95273.pdf}
}

@article{bjerke2023scaling,
  title={Scaling up cell-counting efforts in neuroscience through semi-automated methods},
  author={Bjerke, IE and Yates, SC and Carey, H and Bjaalie, JG and Leergaard, TB},
  journal={iScience},
  volume={26},
  number={12},
  pages={107744},
  year={2023},
  publisher={Cell Press},
  url={https://www.cell.com/iscience/pdf/S2589-0042(23)01639-5.pdf}
}
\end{document}